\documentclass{article} 
\usepackage{iclr2025_conference,times}

\usepackage{amsmath,amsfonts,bm}

\def\eqref#1{equation~\ref{#1}}

\def\1{\bm{1}}

\DeclareMathAlphabet{\mathsfit}{\encodingdefault}{\sfdefault}{m}{sl}
\SetMathAlphabet{\mathsfit}{bold}{\encodingdefault}{\sfdefault}{bx}{n}

\usepackage{hyperref}       
\usepackage{cleveref}
\crefname{figure}{Fig.}{Figs.}
\crefname{table}{Table.}{Tables.}
\crefname{section}{Section}{Secs.}

\usepackage{graphicx}
\usepackage{tabularx}
\usepackage{siunitx}
\usepackage{booktabs}
\usepackage{multirow}
\usepackage{multicol}
\usepackage{makecell}
\usepackage{subcaption}

\usepackage{url}
\usepackage[table]{xcolor}
\definecolor{darkblue}{RGB}{0,0,139}
\definecolor{lightblue}{RGB}{242,249,255}
\definecolor{lightpink}{RGB}{255,239,239}
\definecolor{lightgray}{RGB}{243,243,243}

\usepackage{floatrow}
\newfloatcommand{capbtabbox}{table}[][\FBwidth]
\usepackage{wrapfig}
\usepackage{caption}

\usepackage{pifont}
\usepackage{amssymb}
\newcommand{\tabsize}{\fontsize{7.5pt}{10pt}\selectfont}
\title{TaCo: A Benchmark for Lossless and Lossy Codecs of Heterogeneous Tactile Data}

\author{Zhengxue Cheng$^{1}$\thanks{Corresponding Author} \; Yan Zhao$^{1}$ \; Keyu Wang$^{1}$ \; Hengdi Zhang$^{2}$ \; Li Song$^{1}$  \\
$^{1}$ Shanghai Jiao Tong University, Shanghai, China \\
$^{2}$ Paxini Tech., Shenzhen, China \\
\texttt{zxcheng@sjtu.edu.cn} \\
}

\iclrfinalcopy 
\begin{document}

\maketitle

\begin{figure}[h]
    \centering
    \vspace{-15pt}
    \includegraphics[width=\linewidth]{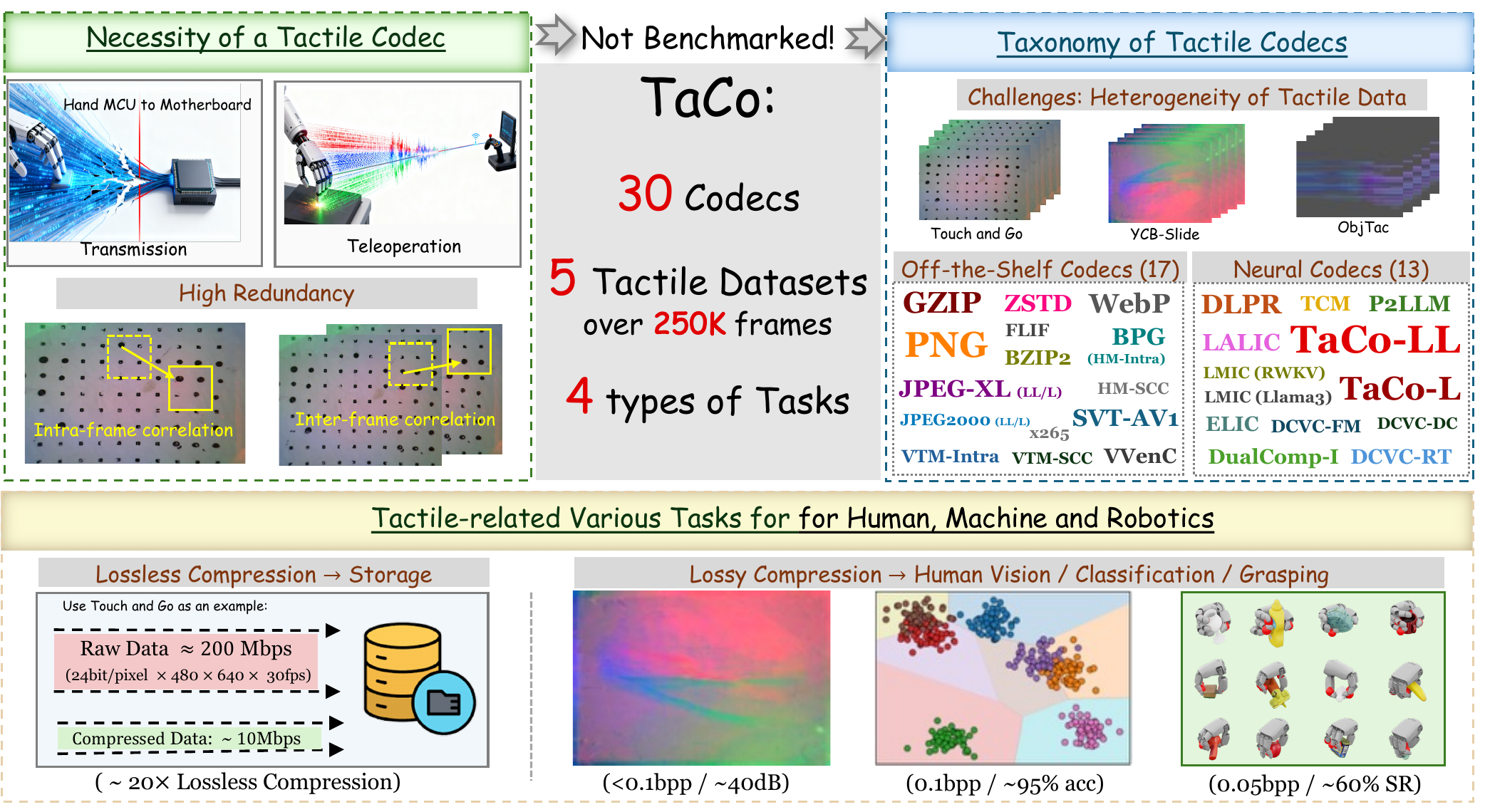}
    \caption{\footnotesize The motivation of our \textbf{TaCo} benchmark, established through an extensive evaluation on tactile codecs across multiple dimensions. First, we assess 30 off-the-shelf and neural codecs on 5 heterogeneous tactile datasets with more than 250K frames. Second, we introduce purely-trained TaCo-LL and TaCo-L codecs to explore the data-driven approaches in the field of lossless and lossy tactile data compression. Finally, we evaluate the coding performance on 4 distinct task types designed to serve for human, machine, and robotics.}
    \label{fig:teaser}
\end{figure}

\begin{abstract}

Tactile sensing is crucial for embodied intelligence, providing fine-grained perception and control in complex environments. However, efficient tactile data compression, which is essential for real-time robotic applications under strict bandwidth constraints, remains underexplored. The inherent heterogeneity and spatiotemporal complexity of tactile data further complicate this challenge. To bridge this gap, we introduce \textbf{TaCo}, the first comprehensive benchmark for \textbf{Ta}ctile data \textbf{Co}decs. TaCo evaluates 30 compression methods, including off-the-shelf compression algorithms and neural codecs, across five diverse datasets from various sensor types. We systematically assess both lossless and lossy compression schemes on four key tasks: lossless storage, human visualization, material and object classification, and dexterous robotic grasping. Notably, we pioneer the development of data-driven codecs explicitly trained on tactile data, \textbf{TaCo-LL} (lossless) and \textbf{TaCo-L} (lossy). Results have validated the superior performance of our TaCo-LL and TaCo-L. This benchmark provides a foundational framework for understanding the critical trade-offs between compression efficiency and task performance, paving the way for future advances in tactile perception.

\end{abstract}
\section{Introduction}

The acquisition and interpretation of tactile data are paramount for advancing embodied AI and achieving sophisticated human-machine interaction, as they provide the rich, physical context necessary for dexterous manipulation and awareness of the environment. However, the high-dimensional, spatio-temporally dense nature of tactile sensing results in rapidly growing data volumes, posing a significant bottleneck for real-time applications. Consequently, efficient tactile data compression is critical for real-time haptic feedback in dexterous hands, remote teleoperation, and large-scale storage of physical interactions for robotic model training.

While the imperative for efficient tactile data compression is well-established, current approaches remain diverse and fragmented. A corpus of existing research has explored this challenge through classical signal processing techniques (like dimensionality reduction and wavelet transforms), leveraging transforms and codecs designed for speech or image data. 
In recent years, data-driven methods have gradually gained popularity for their ability to learn optimal compact representations. Specifically, neural networks can learn compact latent representations in a data-driven manner, enabling efficient lossless or lossy compression \citep{tcm, l3c}. Compared to traditional codecs, neural compression offers greater flexibility and adaptability, particularly in scenarios involving complex or irregular signal structures \citep{ma2019image}. These methods have been successfully applied in domains such as video and image compression \citep{dualcomp, lalic}, but they are still unexplored for tactile data. Another difficulty in generating a data-driven codec for tactile datasets is the heterogeneity, arising from different sensing principles: visuo-tactile sensors such as Gelsight \citep{yuan2017gelsight} and DIGIT \citep{lambeta2020digit} capture surface transformation, while other force-based sensors \citep{paxini2025px6ax} measure force data. 
Therefore, the establishment of a comprehensive and open benchmark, comprising representative datasets, standardized evaluation metrics, and baseline models, is not merely beneficial but a necessary prerequisite for catalyzing advancements in this critical domain and enabling new research.

As illustrated in \cref{fig:teaser}, we construct a comprehensive benchmark to evaluate the compression performance of various codecs on heterogeneous tactile datasets. First, we collect five diverse tactile datasets, and 30 representative codecs. They include off-the-shelf codecs designed for text, image and video, aiming to remove the 1D and 2D, inter-frame and intra-frame redundancies. We also incorporate neural codecs pretrained on other modalities to assess their cross-domain generalization on tactile data. Second, we propose two data-driven codecs, i.e. TaCo-LL and TaCo-L, which are trained from tactile datasets to learn intrinsic data patterns and exploit the redundancies in heterogeneous tactile data. Third, we evaluate tactile compression performance using four types of tasks: lossless compression, lossy compression for human perception, semantic classification, and robot grasping. Experimental results validate the superior performance of our proposed data-driven models, TaCo-LL and TaCo-L, and we hope our benchmark will inspire further research in this field.

In summary, our main contributions are as follows.
\begin{itemize}
    \item We propose \textbf{TaCo}, the first comprehensive benchmark for tactile data codecs. It comprises five publicly tactile datasets, 30 codecs, and four tactile-related tasks: lossless compression, lossy compression for human visualization, tactile classification, and robotic grasping.
    \item We introduce \textbf{TaCo-LL} and \textbf{TaCo-L}, the first purely data-driven tactile codecs, trained end-to-end on heterogeneous tactile datasets to learn the intrinsic data distributions.
    \item Extensive experimental results demonstrate that our proposed TaCo-LL and TaCo-L models surpass existing methods across all four tasks, establishing a foundation for future research in the field of tactile data compression.
\end{itemize}

\section{Related Work}

\paragraph{Tactile Datasets.}

Tactile datasets play a key role in advancing robotic perception and manipulation, supporting tasks like grasping, object recognition, and material classification. Several recent datasets focus only on tactile signals \citep{liu2024tipbench, zhao2024fota, suresh2023ycbslide, yuan2018activecloth, higuera2024tacbench, schneider2025tactilemnist}. TIP Bench \citep{liu2024tipbench} converts sensor outputs into heatmaps and evaluates spatial acuity, stability, and generalization. FoTa \citep{zhao2024fota} merges multiple open datasets into a unified collection of over three million samples. YCB-Slide \citep{suresh2023ycbslide} records both simulated and real sliding interactions between a DIGIT tactile sensor and YCB objects. TacBench \citep{higuera2024tacbench} comprises 180,000+ unlabeled tactile images from surface-sliding interactions, facilitating large-scale self-supervised learning. Tactile MNIST \citep{schneider2025tactilemnist} provides both simulated and real GelSight interactions for MNIST digits, including 13,580 3D-printable meshes and 153,600 tactile recordings. ActiveCloth \citep{yuan2018activecloth} comprises 6,616 robotic squeeze trials on 153 garments, recording synchronized GelSight tactile videos and Kinect depth with 11 attributes.

Beyond pure tactile sensing, some other recent works \citep{yang2022touchandgo, feng2025tacquad-anytouch, liuvtdexmanip, cheng2025touch100k, yu2024PHYSICLEAR, suresh2024feelsight, ssvtp, yuan2017gelfabric, li2019visgel, gao2021objectfolder, cheng2025omnivtla} incorporate multi-modal signals, combining tactile with vision, language, or audio to support cross-modal learning. For instance, Touch and Go \citep{yang2022touchandgo} is the first large-scale tactile dataset collected in outdoor environments, capturing human interactions with natural objects via synchronized tactile and video data. VTDexManip \citep{liuvtdexmanip} provides 565,000 frames of video–tactile data from human multi-finger manipulations across 10 tasks and 182 objects, filling a gap in dexterous interaction datasets. Touch100K \citep{cheng2025touch100k} compiles and cleans TAG and VisGel data into 100,147 high-quality triplets, offering the first large-scale alignment across tactile, visual, and linguistic modalities. TacQuad \citep{feng2025tacquad-anytouch} integrates four visual–tactile sensors, recording aligned tactile signals, RGB frames, and GPT-generated textual descriptions for multimodal reasoning. ObjectForlder \citep{gao2021objectfolder} provides 100 neural object representations that encode 3D shape, appearance, sound, and tactile properties, supporting on-demand multimodal data generation for unified perception and control.

\paragraph{Tactile Compression.}

While tactile sensing continues to advance rapidly in resolution, sampling rate, and coverage, compression research for tactile data remains under-explored. Existing work has proposed some sparse or task-specific compression approaches \citep{hollis2016compressed, bartolozzi2017event, hollis2018compressed, shao2020compression, hassen2020pvc, seeling2021real, liu2023online, slepyan2024wavelet, li2025taccompress, lu2025cross}. For instance, \citet{shao2020compression} exploits the propagation of mechanical waves during dynamic contact to enable compact tactile encoding. \citep{hassen2020pvc} proposes a perceptual vibrotactile codec that combines sparse linear prediction with an acceleration sensitivity function. \citep{seeling2021real} achieves real-time tactile compression by combining bit-level truncation with delta-coding driven by just-noticeable-difference thresholds. Others like \citep{liu2023online} and \citep{slepyan2024wavelet} investigate dimensionality reduction via stacked auto-encoders or wavelet sparsification. However, these methods typically focus on simple signal sparsity or quantization strategies, often lack rigorous compression metrics and are tailored to relatively narrow scenarios or limited generalizability. In fact, many common tactile signals can be naturally transformed into image-like formats, enabling the use of standard image or general-purpose compressors. This direction is appealing not only because these compressors are well-established and widely available, but also because they offer tunable configurations to trade off compression ratio and distortion, making them adaptable to diverse robotic applications. Yet, this perspective remains largely under-explored in the tactile domain. To fill this gap, this paper presents a comprehensive benchmark for tactile compression methods, aiming to provide practical guidance and spark future research into efficient tactile data compression.


\section{Tactile Datasets and Compression Methods}
\label{sec:datasets}

\subsection{Tactile Datasets}
\label{sec:datasets}

We benchmark tactile compression across five representative datasets: Touch and Go \citep{yang2022touchandgo}, ObjectFolder 1.0 \citep{gao2021objectfolder}, SSVTP \citep{ssvtp}, YCB-Slide \citep{suresh2023ycbslide}, and ObjTac \citep{cheng2025omnivtla}. These datasets span a range of sensor types (vision-based and force-based), resolutions (from $120 \times 160$ to $640 \times 480$), and data scales, as detailed in \cref{tab:datasets}. Depending on the sensor type, tactile data exhibit strong structural heterogeneity, along with complex spatiotemporal correlations and redundancy, as illustrated in \cref{fig:teaser}.

Specifically, the GelSight-based datasets (Touch and Go and ObjectFolder) and DIGIT-based datasets (SSVTP and YCB-Slide) are collected using vision-based tactile sensors that operate by illuminating a deformable elastomer surface with micro-LED arrays and capturing its surface deformation through an internal camera. This process converts tactile interactions into sequences of RGB images or videos, enabling direct applications of image or video compression techniques. In contrast, the ObjTac dataset is collected using force-based tactile sensors. The sensor comprises $N=60$ contact points across the contact surface, each measuring a 3D force vector. These measurements form a temporally structured sequence of force data. To enable efficient compression, we map each 3D force vector to an RGB pixel and temporally stack the force readings across a time duration $T$, generating images of resolution $T \times 60$.

\begin{table}[!tb]
    \centering
    \footnotesize
    \belowrulesep=0pt
    \aboverulesep=0pt
    \renewcommand{\arraystretch}{1.2}
    \setlength{\tabcolsep}{2pt}
    \begin{tabular}{l|llll} 
    \toprule
    \textbf{Dataset} 
    & \textbf{\#Objects} 
    & \textbf{\#Frames} 
    & \textbf{Resolution}
    & \textbf{Sensor} 
    \\
    \midrule
    Touch and Go \citep{yang2022touchandgo} 
    & 3971 & 13.9K & $640\times480 \times 30$Hz 
    & GelSight \citep{yuan2017gelsight} \\
    ObjectFolder 1.0 \citep{gao2021objectfolder}
    & 100 & 100K & $120\times160\times 30$Hz
    & GelSight \citep{yuan2017gelsight} \\
    SSVTP \citep{ssvtp}
    & 10 & 4.5K & $240\times320\times 30$Hz
    & DIGIT \citep{lambeta2020digit} \\
    YCB-Slide \cite{suresh2023ycbslide}
    & 10 & 4.5K & $240\times320\times 30$Hz 
    & DIGIT \citep{lambeta2020digit}\\
    ObjTac \citep{cheng2025omnivtla} 
    & 56 & 135K & $5\times12 \times 200$Hz
    & Force Sensor \citep{paxini2025px6ax}\\
    \bottomrule
    \end{tabular}
    \caption{\footnotesize Introduction of the utilized tactile datasets.}
    \label{tab:datasets}
\end{table}

\subsection{Tactile Compression Methods}
\label{sec:methods}

We establish a benchmark for two categories of tactile codecs: 1) \textit{off-the-shelf codecs} based on conventional signal processing, originally designed for general-purpose or visual data, and 2) \textit{neural codecs} that leverage neural networks to learn data patterns end-to-end. 
As tactile signals are natively or transformable into image or video formats (see \cref{sec:datasets}), our evaluation of neural codecs includes both pretrained image codecs and, to our knowledge, the first data-driven codecs explicitly trained on tactile datasets.

\subsubsection{Off-the-Shelf Compression Methods}

Typically, off-the-shelf compression methods have been historically designed for text, image and video data. These classical techniques are fundamentally rooted in signal processing principles, aiming to eliminate statistical, spatial, or temporal redundancies present in 1D or 2D data.

\paragraph{General-Purpose Compression Methods.}

We evaluate three general-purpose lossless compressors: gzip \citep{gzip}, zstd \citep{zstd}, and bzip2 \citep{bzip2}, which are designed to exploit 1D symbol redundancy using techniques such as dictionary coding (e.g., LZ77 in gzip and zstd), block-sorting transforms (e.g., Burrows-Wheeler in bzip2), and entropy coding. 

\paragraph{Image and Video Compression Methods.}

When treating tactile data as images, we evaluate six standard image lossless codecs: PNG \citep{png}, FLIF \citep{flif}, WebP \citep{webp}, JPEG-XL \citep{jpegxl}, JPEG2000 \citep{jpeg2000}, and BPG \citep{bpg} (the intra-mode of HEVC/265 codec). These image-specific compressors remove 2D spatial redundancy in images via predictive coding, transform coding (e.g., DCT or wavelets), and context-based entropy coding.

In addition, we also evaluate six lossy image codecs: JPEG-XL \citep{jpegxl}, JPEG2000 \citep{jpeg2000}, as well as the intra-frame and screen content coding (SCC) modes of HM \cite{hevc} and VTM \citep{vvc} (i.e., HM-Intra, HM-SCC, VTM-Intra, VTM-SCC). 

To further address the inter-frame redundancy in tactile data, we evaluate three off-the-shell video codecs, VVenC \citep{vvc}, x265 \citep{hevc}, and SVT-AV1 \citep{av1}. Due to the huge amount of video data, lossless video compression is rarely used in practice, so we only discuss lossy video compression methods.

\subsubsection{Neural Compression Methods}

\begin{figure}[tb]
    \centering
    \includegraphics[width=0.96\linewidth]{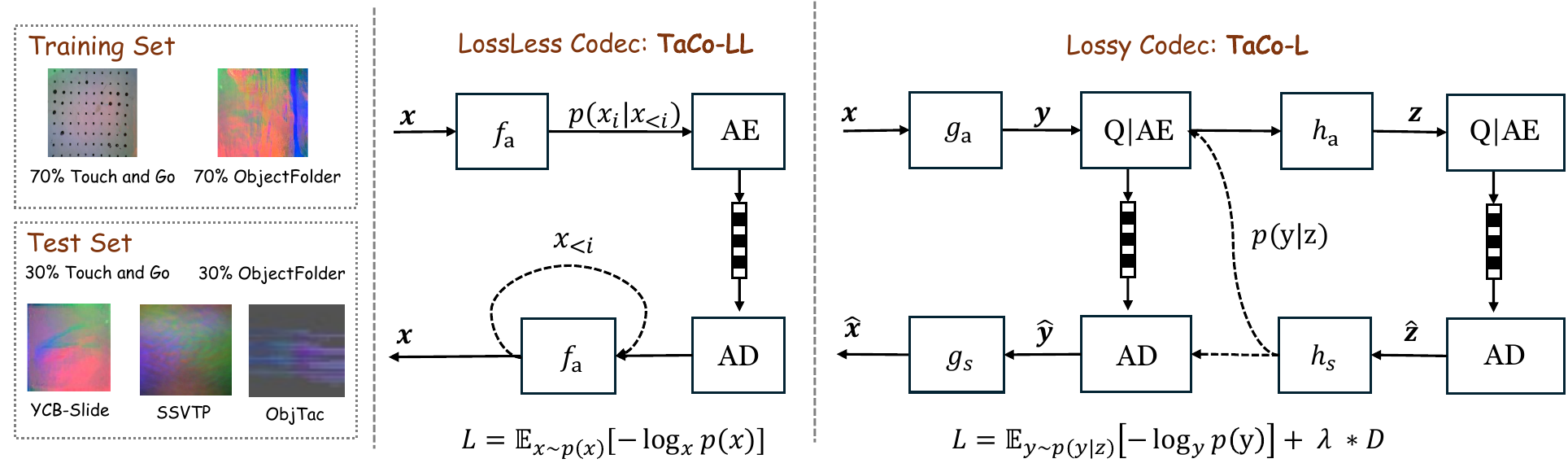}
    \caption{\footnotesize Diagram of data-driven compression methods and our proposed TaCo-L and TaCo-LL. }
    \vspace{-2mm}
    \label{fig:overview}
\end{figure}

Recently, neural codecs have surpassed conventional codecs on text, image and video, owing to powerful learning capabilities of neural networks to fit the latent data distribution. However, tactile data exhibit unique statistical patterns, and heterogeneous tactile datasets involve different distributions, potentially making pre-trained neural encoders less applicable.
Herein, we briefly introduce the diagram of learning-based lossy and lossless neural codecs, as ~\cref{fig:overview}. 

In the case of lossless compression, the tactile signal $\boldsymbol{x}$ is sequentially fed into a neural network $f_{a}$ to predict the distribution of next symbol, $p(x_i|x_{<i})$, then it is followed by an arithmetic encoder (AE) to generate bitstream. The loss is the entropy, which is the minimal bound to encode $\boldsymbol{x}$:
\begin{equation}
    \mathcal{L} = \mathop{\mathbb{E}}[-\log_{2}(p(x_i|x_{<i}))]
\end{equation}
At the decoder side, the symbols can be lossless autoregressively decoded through an arithmetic decoder (AD) and the same network. 

For neural lossy codecs, the tactile signal $\boldsymbol{x}$ is transformed through a transform function $g_{a}$ into a latent presentation $\boldsymbol{y}$. Afterwards, $\boldsymbol{y}$ is quantized through $Q$ to get discrete values $\hat{\boldsymbol{y}}$, then it is followed by AE to generate bitstream. At the decoder side, $\hat{\boldsymbol{y}}$ is decoded from bitstream using an AD and then transformed back to reconstructed images $\hat{\boldsymbol{x}}$ though an inverse transform function $g_{s}$. $h_{a}$ and $h_{s}$ denote analysis and synthesis transforms in the hyper autoencoder to generate side bits $\boldsymbol{z}$, as a prior to estimate density model of $\boldsymbol{\hat{y}}$. The loss is defined as a rate-distortion function:
\begin{equation}
    \mathcal{L} = \lambda\times\mathcal{D} (\boldsymbol{x}, \boldsymbol{\hat{x}}) + \mathop{\mathbb{E}}[-\log_{2}(p_{\hat{\boldsymbol{y}}|\hat{\boldsymbol{z}}}(\hat{\boldsymbol{y}}|\hat{\boldsymbol{z}}  ))] 
\end{equation}
where $\lambda$ is a hyper-parameter to control the bitrate, and all the parameters are learnable.

\paragraph{Pretrained Neural Codecs}

The lossless neural-based methods include 5 image compression models, DLPR \citep{dlpr}, P2LLM \citep{p2llm}, and DualComp-I \citep{dualcomp}, as well as LMIC \citep{lmic}, a multi-modality compressor based on pretrained large language models (specifially, in this paper we use RWKV-7B \citep{rwkv7} and Llama3-8B \citep{llama3} as LLMs for implementation). 
It is worth noting that these models are pretrained on natural language or image data, and evaluated on tactile data without any domain-specific adaptations. 

For the purpose of lossy neural-based compression approaches, we evaluate a total of 6 compression models, consisting of three recent neural-based image codecs (ELIC \citep{elic}, TCM \citep{tcm}) and LALIC \citep{lalic}, and three recent neural-based video compressors (DCVC-DC \citep{dcvc-dc}, DCVC-FM \citep{dcvc-fm}, and DCVC-RT \citep{dcvc-rt}).

\paragraph{Tactile Data-Driven Neural Codecs}

To our knowledge, there have been yet no existing methods fully trained on tactile signals to explore the upper bound of tactile compression performance. 
To further explore the performance potential of data-driven codec, we retrain two state-of-the-art compression models, DualComp-I and LALIC, using tactile datasets. 
Specifically, DualComp-I operates lossless compression by tokenizing the input into discrete representations and applying an auto-regressive model to predict each token's distributions, enabling efficient entropy coding. LALIC implements lossy compression and follows a variational auto-encoder (VAE) \citep{vae} architecture. The two models are chosen for their competitive performance and efficiency in their respective domains. By retraining them on tactile data, we aim to assess the benefits of data-driven tactile compression. The retrained models are referred to as \textbf{TaCo-LL} (lossless) and \textbf{TaCo-L} (lossy), respectively, to distinguish them from their original pretrained versions.

\begin{figure*}[tb]
    \centering
    \includegraphics[width=0.9\linewidth]{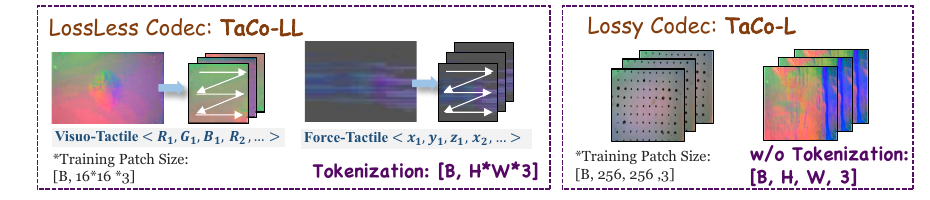}
    \caption{\footnotesize Detailed implementations of our proposed TaCo-L and TaCo-LL.} 
    \vspace{-2mm}
    \label{fig:token}
\end{figure*}

Specifically, for \textbf{TaCo-LL}, the tokenization is conducted as shown in~\cref{fig:token}. We divide the input into $16\times16\times3$ patches to preserve local spatial correlations. We then flatten the data in a raster-scan order. For visuo-tactile data, including Touch and Go, YCB-slide, ObjectFolder, SSVTP, the RGB values are sequentially expanded as sub-pixels $(R_1, G_1, B_1, R_2, G_2, B_2, \cdot\cdot\cdot)$. For three-axis force signals, i.e. ObjTac, are treated as three color channels and expanded as $(x_1, y_1, z_1, x_2, y_2, z_2, \cdot\cdot\cdot)$.
For \textbf{TaCo-L}, we follow the setup of LALIC~\footnote{\url{https://github.com/sjtu-medialab/RwkvCompress}} and randomly crop or zero-pad the input tactile image to $256\times256$ resolution. Since the input tensor has three channels for both visuo-tactile data and force-tactile data, no tokenization is needed, as shown in ~\cref{fig:token}. The network architecture is adopted from the LALIC model~\citep{lalic} and the $g_a$ and $g_s$ consist of four downsampling and upsampling operations, respectively.

To this end, we benchmark a total of \textbf{30 codecs} to evaluate the compressibility of tactile data. Among them, 14 codecs (9 off-the-shell codecs, 4 neural codecs and one proposed TaCo-LL) support lossless compression, aiming to preserve exact signal fidelity. The remaining codecs (9 off-the-shell, 6 neural codecs and one proposed TaCo-L) are lossy, targeting higher compression ratios at the cost of some reconstruction distortion. 
These methods can also be categorized by their training data domain: 28 codecs are existing methods originally developed for general-purpose or visual data and applied without any tactile-specific adaptation, while the remaining two (TaCo-LL, TaCo-L) are data-driven models explicitly trained on tactile datasets. Evaluating the pretrained models allows us to assess how well existing compression techniques generalize to tactile data, whereas the tactile data-driven methods help explore the potential of tactile-aware compression strategies.

\section{Experiments}

\subsection{Experimental Setup}

We benchmark the performance of tactile compression on five representative tactile datasets: Touch and Go, ObjectFolder 1.0, SSVTP, YCB-Slide, and ObjTac, as shown in \cref{sec:datasets}. 
Specifically, we randomly select 70\% of the data from the Touch and Go and ObjectFolder datasets to train the TaCo-LL and TaCo-L models. The remaining 30\% of the two datasets, together with the entire SSVTP, YCB-Slide and ObjTac datasets, are used for all methods' compression evaluation. Following \citep{dualcomp}, for TaCo-LL we use the FusedAdam optimizer \citep{fusedadam} with a cosine annealing learning rate schedule \cite{scheduler}, starting at $1\times 10^{-4}$ and decaying to $2\times 10^{-5}$ over 20 epochs. Following \citep{lalic}, we train TaCo-L using the Adam optimizer \citep{adam}. The learning rate is set to $1\times 10^{-4}$ for 40 epochs, then decayed to $1\times 10^{-5}$ for another 4 epochs. The training is performed on two NVIDIA A100 GPUs.

\begin{table*}[tb]
    \belowrulesep=0pt
    \aboverulesep=0pt
    \centering
    \caption{\footnotesize Comparison of lossless compression performance (bits/Byte) on five tactile datasets. The best results are highlighted in \textbf{\color{darkblue}bold blue}, second-best in \textbf{bold}, and third to fifth in \underline{underline}. For TaCo, 12M/48M/96M denotes the model parameter. To show the compression performance more clearly, we also list the compression ratios relative to the uncompressed data (8 bits/Byte) in parentheses only for the best and second best results.}
    \tabsize
    \renewcommand{\arraystretch}{1.1}
    \setlength{\tabcolsep}{6.5pt}
    \newcommand{\centerdash}[1]{\ifx#1-\multicolumn{1}{c}{-}\else#1\fi}
    \vspace{-6pt}
    \begin{tabular}{p{0.24cm}p{4cm}|lllll}
    \toprule
     & \multirow{2}{*}{\textbf{Compressor}} 
     & \multicolumn{5}{c}{\textbf{bits/Byte$\downarrow$}}\\[1pt]
     \cmidrule{3-7}

     & 
     & TouchandGo
     & ObjectFolder
     & SSVTP
     & ObjTac
     & YCB-Slide\\
    \midrule

    & \cellcolor{lightgray}{uncompressed}        
    & \cellcolor{lightgray}{8 ($1\times$)}
    & \cellcolor{lightgray}{8 ($1\times$)} 
    & \cellcolor{lightgray}{8 ($1\times$)} 
    & \cellcolor{lightgray}{8 ($1\times$)} 
    & \cellcolor{lightgray}{8 ($1\times$)}\\
    \cmidrule{1-7}
    \multirow{9.3}{*}{\rotatebox{90}{\textbf{Off-the-Shelf}}} 
    & gzip \citep{gzip}   & 2.298 & 3.969 & 2.234 &  0.571 & 2.185\\
    & zstd \citep{zstd}   & 2.263 & 3.966 & 2.233 &  0.568 & 2.184\\
    & bzip2 \citep{bzip2} & 2.288 & 4.031 & 2.255 &  0.594 & 2.205\\
    & FLIF \citep{flif}   & \underline{0.808} ($10\times$) & 3.765 & 1.567 &  \textbf{0.363} ($22\times$) & \underline{1.489} ($5\times$)\\
    & BPG  \citep{bpg}    & 1.293 & 3.726 & 2.000 &  0.513 & 1.922\\
    & WebP \citep{webp}   & 0.936 & 3.612 & 1.820 &  \underline{0.424} ($19\times$) & 1.767\\
    & JPEG-XL \citep{jpegxl}    & \underline{0.739} ($11\times$) & 3.657 & \underline{1.478} ($5\times$) & \underline{0.382} ($21\times$) & \underline{1.431} ($6\times$)\\
    & JPEG2000 \citep{jpeg2000} & 1.552 & 3.989 & 1.997 & 1.399 & 1.916\\
    & PNG \citep{png}    & 2.500 & 3.964 & 2.233 & 0.579 & 2.183\\
    \cmidrule{1-7}

    \multirow{8.3}{*}{\rotatebox{90}{\textbf{Neural}}}
    & DLPR \citep{dlpr} & 1.082 & 3.774 & 1.539 & 0.522 & 1.503\\
    & P2LLM \citep{p2llm}  & 1.212 & \underline{3.400} & 1.804 &  0.546 & 1.512\\
    & Llama3* \citep{lmic} & 2.055 & 3.465 & 1.975 &  0.834 & 1.905\\
    & RWKV* \citep{lmic}   & 2.223 & 3.718 & 2.010 &  0.540 & 1.880\\
    & DualComp-I \citep{dualcomp} & 0.948 & \underline{3.126} ($3\times$) & \underline{1.442} ($6\times$) & 0.540 & \underline{1.388} ($6\times$)\\

    & \cellcolor{lightpink}TaCo-LL-12M (ours)
    & \cellcolor{lightpink}\underline{0.622} ($13\times$)
    & \cellcolor{lightpink}\underline{3.098} ($3\times$)
    & \cellcolor{lightpink}\underline{1.457} ($6\times$)
    & \cellcolor{lightpink}0.569
    & \cellcolor{lightpink}1.520\\
    
    & \cellcolor{lightpink}TaCo-LL-48M (ours) 
    & \cellcolor{lightpink}\textbf{0.504 ($16\times$)}
    & \cellcolor{lightpink}\textbf{2.923 ($3\times$)}
    & \cellcolor{lightpink}\textbf{1.249 ($6\times$)}
    & \cellcolor{lightpink}\underline{0.411} ($20\times$)
    & \cellcolor{lightpink}\textbf{1.321 ($6\times$)}\\
    
    & \cellcolor{lightpink}TaCo-LL-96M (ours) 
    & \cellcolor{lightpink}\textcolor{darkblue}{\textbf{0.447} ($18\times$)}  
    & \cellcolor{lightpink}\textcolor{darkblue}{\textbf{2.709} ($3\times$)}  
    & \cellcolor{lightpink}\textcolor{darkblue}{\textbf{1.066} ($8\times$)} 
    & \cellcolor{lightpink}\textcolor{darkblue}{\textbf{0.360} ($22\times$)}
    & \cellcolor{lightpink}\textcolor{darkblue}{\textbf{1.073} ($8\times$)}\\
    \bottomrule    
    \end{tabular}
    \label{tab:compare-lossless-rd}
\end{table*}

\begin{table*}[tb]
    \belowrulesep=0pt
    \aboverulesep=0pt
    \centering
    \tabsize
    \caption{\footnotesize The complexity of lossless compression algorithms on five tactile datasets. $\dag$ and $\ddag$ indicates speeds measured on MacBook Pro CPU and NVIDIA A100 GPU, respectively.}
    \renewcommand{\arraystretch}{1.1}
    \setlength{\tabcolsep}{2.5pt}
    \newcommand{\centerdash}[1]{\ifx#1-\multicolumn{1}{c}{-}\else#1\fi}
    \vspace{-6pt}
    \begin{tabular}{p{0.22cm}p{3.5cm}|lll lllll}
    \toprule
     & \multirow{2}{*}{\textbf{Compressor}} 
     & \multirow{2}{*}{\textbf{\#Params}$\downarrow$} 
     & \multirow{2}{*}{\textbf{MACs}$\downarrow$} 
     & \multirow{2}{*}{\textbf{\makecell{Speed\\(KB/s)}}$\uparrow$}
     & \multicolumn{5}{c}{\textbf{Speed (FPS)$\uparrow$}}  \\ [1pt]
     \cmidrule{6-10}
     
     & 
     & 
     & 
     & 
     & TouchandGo
     & ObjectFolder
     & SSVTP
     & ObjTac
     & YCB-Slide\\
    
    \midrule

    
    \midrule
    \multirow{9.3}{*}{\rotatebox{90}{\textbf{Off-the-Shelf}}} 
    & gzip \citep{gzip}   & - & - & 14500$^\dag$
    & 15.7$^\dag$ & 252$^\dag$ & 62.9$^\dag$ & 190$^\dag$ & 63.9$^\dag$ \\
    & zstd \citep{zstd}   & - & - & 11000$^\dag$
    & 11.9$^\dag$ & 191$^\dag$ & 47.7$^\dag$ & 144$^\dag$ & 47.4$^\dag$ \\
    & bzip2 \citep{bzip2} & - & - & 3300$^\dag$
    & 3.58$^\dag$ & 57.3$^\dag$ & 14.3$^\dag$ & 43.3$^\dag$ & 14.3$^\dag$ \\
    & FLIF \citep{flif}   & - & - & 652$^\dag$
    & 0.71$^\dag$ & 11.3$^\dag$ & 2.84$^\dag$ & 8.56$^\dag$ & 2.84$^\dag$ \\
    & BPG  \citep{bpg}    & - & - & 180$^\dag$
    & 0.20$^\dag$ & 3.13$^\dag$ & 0.78$^\dag$ & 2.36$^\dag$ & 0.78$^\dag$ \\
    & WebP \citep{webp}   & - & - & 330$^\dag$
    & 0.36$^\dag$ & 5.73$^\dag$ & 1.43$^\dag$ & 4.33$^\dag$ & 1.43$^\dag$ \\
    & JPEG-XL \citep{jpegxl}  & - & - & 970$^\dag$ 
    & 1.05$^\dag$ & 16.8$^\dag$ & 4.21$^\dag$ & 12.7$^\dag$ & 4.21$^\dag$ \\
    & JPEG2000 \citep{jpeg2000} & - & - & 5000$^\dag$
    & 5.43$^\dag$ & 86.8$^\dag$ & 21.7$^\dag$ & 65.7$^\dag$ & 21.7$^\dag$ \\
    & PNG \citep{png}   & - & - & 200$^\dag$
    & 0.22$^\dag$ & 3.47$^\dag$ & 0.87$^\dag$ & 2.63$^\dag$ & 0.87$^\dag$ \\
    \midrule

    \multirow{8.3}{*}{\rotatebox{90}{\textbf{Neural}}}
    & DLPR \citep{dlpr} & 22.3M & - & 640$^\ddag$
    & 0.69$^\ddag$ & 11.1$^\ddag$ & 2.78$^\ddag$ & 8.41$^\ddag$ & 2.78$^\ddag$ \\
    & P2LLM \citep{p2llm}  & 8B & - & 20$^\ddag$
    & 0.02$^\ddag$ & 0.35$^\ddag$ & 0.09$^\ddag$ & 0.26$^\ddag$ & 0.09$^\ddag$ \\
    & Llama3* \citep{lmic} & 8B & 7.8G & 20$^\ddag$ 
    & 0.02$^\ddag$ & 0.35$^\ddag$ & 0.09$^\ddag$ & 0.26$^\ddag$ & 0.09$^\ddag$ \\
    & RWKV* \citep{lmic}   & 7B & 7.2G & 86$^\ddag$
    & 0.09$^\ddag$ & 1.49$^\ddag$ & 0.37$^\ddag$ & 1.13$^\ddag$ & 0.37$^\ddag$ \\
    & DualComp-I \citep{dualcomp} & 96M & 59.9M & 317$^\ddag$
    & 0.34$^\ddag$ & 5.50$^\ddag$ & 1.38$^\ddag$ & 4.16$^\ddag$ & 1.38$^\ddag$ \\

    & \cellcolor{lightpink}TaCo-LL-12M (ours)
    & \cellcolor{lightpink}12M 
    & \cellcolor{lightpink}11.6M 
    & \cellcolor{lightpink}614$^\ddag$
    & \cellcolor{lightpink}0.67$^\ddag$
    & \cellcolor{lightpink}10.7$^\ddag$
    & \cellcolor{lightpink}2.66$^\ddag$
    & \cellcolor{lightpink}8.06$^\ddag$
    & \cellcolor{lightpink}2.66$^\ddag$ \\
    
    & \cellcolor{lightpink}TaCo-LL-48M (ours) 
    & \cellcolor{lightpink}48M 
    & \cellcolor{lightpink}33.3M 
    & \cellcolor{lightpink}360$^\ddag$
    & \cellcolor{lightpink}0.39$^\ddag$ 
    & \cellcolor{lightpink}6.25$^\ddag$
    & \cellcolor{lightpink}1.56$^\ddag$
    & \cellcolor{lightpink}4.73$^\ddag$
    & \cellcolor{lightpink}1.56$^\ddag$ \\
    
    & \cellcolor{lightpink}TaCo-LL-96M (ours) 
    & \cellcolor{lightpink}96M 
    & \cellcolor{lightpink}59.9M
    & \cellcolor{lightpink}317$^\ddag$
    & \cellcolor{lightpink}0.34$^\ddag$ 
    & \cellcolor{lightpink}5.50$^\ddag$
    & \cellcolor{lightpink}1.38$^\ddag$ 
    & \cellcolor{lightpink}4.16$^\ddag$ 
    & \cellcolor{lightpink}1.38$^\ddag$ \\
    \bottomrule    
    \end{tabular}
    \label{tab:compare-lossless}
\end{table*}

\subsection{Lossless Compression}



\paragraph{Evaluation Metrics.}

We evaluate lossless compression efficiency using bits per Byte, which quantifies the number of bits required to encode one byte of the original tactile data. Lower bits/Byte values indicate more effective compression, with uncompressed data corresponding to 8 bits/Byte. 
We also compare the complexity of different algorithms using four metrics, i.e. model parameters, MACs, inference speed (KB/s) on multiple devices (including NVIDIA A100 GPU, a MacBook Pro), and the frame per second (FPS) ranging with different spatial resolutions.

\begin{table*}[tb]
    \belowrulesep=0pt
    \aboverulesep=0pt
    \centering
    \caption{\footnotesize Evaluation of lossy compression performance on five tactile datasets leveraging intra-frame compressors. The best results are shown in \textbf{\textcolor{darkblue}{blue bold}}, the second-best in \textbf{bold}, and the third-best in \underline{underline}. For the reference, the bandwidth consumption of the anchor HEVC-intra is approximately 2Mbps at the quality of 40dB, which is calculated by $0.22$ bit per pixel $\times 640 \times 480 \times 30$fps$ \times 10^{-6}$ for Touch and Go dataset, as \cref{fig:lossy-image-touchandgo}.}
    \tabsize
    \renewcommand{\arraystretch}{1.1}
    \setlength{\tabcolsep}{8.5pt}
    \newcommand{\centerdash}[1]{\ifx#1-\multicolumn{1}{c}{-}\else#1\fi}
    \vspace{-6pt}
    \begin{tabular}{p{0.3cm}p{3.8cm}|lllll}
    \toprule
     & \multirow{2}{*}{\textbf{Compressor}} 
     & \multicolumn{5}{c}{\textbf{BD-Rate (\%)}$\downarrow$}\\[1pt]
     \cmidrule{3-7}

     & 
     & TouchandGo
     & ObjectFolder
     & SSVTP
     & YCB-Slide
     & ObjTac \\
    \midrule

    \multirow{6.3}{*}{\rotatebox{90}{\textbf{Off-the-Shelf}}} 
    & HM-Intra \citep{hevc}   
    & 0\% & 0\% & 0\% & 0\% & 0\% \\
    & HM-SCC \citep{hevc}   
    & -10.4\% & 2.0\% & 6.9\% & 7.2\%  & \textcolor{darkblue}{\textbf{-44.5\%}} \\
    & VTM-Intra \citep{vvc} 
    & -21.7\% & \textbf{-19.7\%} & \textbf{-16.0\%} & \textbf{-24.4\%} & -22.0\% \\
    & VTM-SCC \citep{vvc} 
    & -23.7\% & \underline{-18.0\%} & \underline{-13.7\%} & \underline{-19.1\%} & \textbf{-44.3\%} \\
    & JPEG-XL \citep{jpegxl}
    & 66.7\% & 60.6\% & 77.5\%  & 96.9\% & 99.4\% \\
    & JPEG2000 \citep{jpeg2000}
    & 59.7\% & 69.9\% & 107.9\% & 89.1\% &103.8\% \\
    \cmidrule{1-7}

    \multirow{4.3}{*}{\rotatebox{90}{\textbf{Neural}}}
    & ELIC \citep{elic}   
    &  \underline{-40.2\%} & 0.6\%  & -5.8\% & -9.2\% & 44.5\%  \\
    & LALIC \citep{lalic} 
    &  \textbf{-51.6\%} & 0.2\%  & 4.3\%  & -4.6\% & 32.8\% \\
    & TCM \citep{tcm}     
    &  -39.9\% & 23.7\% & 42.9\% & 30.5\%  & 97.2\% \\
    
    & \cellcolor{lightblue}TaCo-L (Ours)
    & \cellcolor{lightblue}\textcolor{darkblue}{\textbf{-61.8\%}}
    & \cellcolor{lightblue}\textcolor{darkblue}{\textbf{-24.3\%}}
    & \cellcolor{lightblue}\textcolor{darkblue}{\textbf{-19.2\%}}
    & \cellcolor{lightblue}\textcolor{darkblue}{\textbf{-27.4\%}}
    & \cellcolor{lightblue}\underline{-27.0\%} \\

    \bottomrule    
    \end{tabular}
    \label{tab:compare-lossy-rd}
\end{table*}

\begin{table*}[tb]
    \belowrulesep=0pt
    \aboverulesep=0pt
    \centering
    \caption{\footnotesize The complexity of lossy compression algorithms on five tactile datasets leveraging intra-frame compressors. $\dag$ and $\ddag$ indicates speeds measured on MacBook Pro CPU and NVIDIA A100 GPU, respectively.}
    \tabsize
    \renewcommand{\arraystretch}{1.22}
    \setlength{\tabcolsep}{2.3pt}
    \newcommand{\centerdash}[1]{\ifx#1-\multicolumn{1}{c}{-}\else#1\fi}
    \vspace{-6pt}
    \begin{tabular}{p{0.28cm}p{3.6cm}|lll lllll}
    \toprule
     & \multirow{2}{*}{\textbf{Compressor}} 
     & \multirow{2}{*}{\textbf{\#Params}$\downarrow$} 
     & \multirow{2}{*}{\textbf{MACs}$\downarrow$} 
     & \multirow{2}{*}{\textbf{\makecell{Speed\\(KB/s)}}$\uparrow$}
     & \multicolumn{5}{c}{\textbf{Speed (FPS)$\uparrow$}}  \\ [1pt]
     \cmidrule{6-10}

     & & & &
     & TouchandGo
     & ObjectFolder
     & SSVTP
     & YCB-Slide
     & ObjTac \\
    \midrule

    \multirow{6.3}{*}{\rotatebox{90}{\textbf{Off-the-Shelf}}} 
    & HM-Intra \citep{hevc}  & - & - & 11.1$^\dag$
    & 0.12$^\dag$ & 1.97$^\dag$ & 0.50$^\dag$ & 1.49$^\dag$ & 0.50$^\dag$  \\
    & HM-SCC \citep{hevc}   & - & - & 31.3$^\dag$
    & 0.34$^\dag$ & 0.56$^\dag$ & 0.41$^\dag$ & 0.42$^\dag$ & 0.41$^\dag$ \\
    & VTM-Intra \citep{vvc} & - & - & 9.22$^\dag$
    & 0.10$^\dag$ & 1.63$^\dag$ & 0.41$^\dag$ & 1.23$^\dag$ & 0.41$^\dag$\\
    & VTM-SCC \citep{vvc}   & - & - & 4.61$^\dag$
    & 0.05$^\dag$ & 0.72$^\dag$ & 0.18$^\dag$ & 0.55$^\dag$ & 0.18$^\dag$ \\
    & JPEG-XL \citep{jpegxl} & - & - & 2305$^\dag$
    & 2.50$^\dag$ & 40.0$^\dag$ & 10.0$^\dag$ & 30.2$^\dag$ & 10.0$^\dag$ \\
    & JPEG2000 \citep{jpeg2000} & - & - & 13200$^\dag$
    & 14.3$^\dag$ & 228$^\dag$ & 57.1$^\dag$ & 172$^\dag$ & 57.1$^\dag$ \\
    \midrule

    \multirow{4.3}{*}{\rotatebox{90}{\textbf{Neural}}}
    & ELIC \citep{elic}  & 33.3M & 0.9M & 4075$^\ddag$
    & 4.42$^\ddag$ & 70.7$^\ddag$ & 17.7$^\ddag$ & 53.5$^\ddag$ & 17.7$^\ddag$ \\
    & LALIC \citep{lalic} & 63.2M & 0.7M & 3700$^\ddag$
    & 4.01$^\ddag$ & 64.2$^\ddag$ & 16.1$^\ddag$ & 48.6$^\ddag$ & 16.1$^\ddag$ \\
    & TCM \citep{tcm} & 75.9M & 1.8M & 5680$^\ddag$
    & 6.16$^\ddag$ & 98.6$^\ddag$ & 24.7$^\ddag$ & 74.6$^\ddag$ & 24.7$^\ddag$ \\
    
    & \cellcolor{lightblue}TaCo-L (Ours)
    & \cellcolor{lightblue}63.2M
    & \cellcolor{lightblue}0.73M
    & \cellcolor{lightblue}3700$^\ddag$
    & \cellcolor{lightblue}4.01$^\ddag$
    & \cellcolor{lightblue}64.2$^\ddag$
    & \cellcolor{lightblue}16.1$^\ddag$
    & \cellcolor{lightblue}48.6$^\ddag$
    & \cellcolor{lightblue}16.1$^\ddag$ \\

    \bottomrule    
    \end{tabular}
    \label{tab:compare-lossy}
\end{table*}



\paragraph{Results.}

\cref{tab:compare-lossless-rd} benchmarks the lossless compression performance across five tactile datasets. As expected, all methods obviously reduce the data's storage cost, but the degree of compression varies across the compressors and datasets. Among off-the-shelf baselines, general-purpose compressors such as gzip and zstd achieve moderate compression ratios. Image-specific codecs like FLIF and JPEG-XL provide notably better results especially on vision-like tactile data, due to their ability to exploit spatial correlations. Learning-based methods pretrained on natural images, such as DLPR, P2LLM, and DualComp-I, effectively capture intra-frame correlations in tactile signals and generally provide pleasing results. However, their performance remains limited by domain mismatch, especially on non-visual or structurally different tactile datasets.

To further explore the potential of data-driven compression, we retrain state-of-the-art lossless image compressor, DualComp-I, on tactile datasets and obtain our TaCo-LL model. The largest variant, TaCo-LL-96M, achieves the best performance across all five datasets, reaching 0.447 bits/Byte on TouchandGo, 2.709 bits/Byte on ObjectFolder, 1.066 on SSVTP, 0.360 bits/Byte on ObjTac, and 1.073 on TCB-Slide, corresponding to $18\times$, $3\times$, $8\times$, $22\times$, and $8\times$ compression ratios, respectively. 
\cref{tab:compare-lossless} benchmarks the complexity of different compression algorithms. Off-the-shelf codecs can achieve relatively fast speed. Among neural codecs, TaCo-LL models achieve competitive compression performance with fewer parameters compared to P2LLM, Llama3-8B and RWKV-7B, and the encoding/decoding speed ranges from 317KB/s to 614KB/s. 



\subsection{Lossy Compression for Human Vision}



\paragraph{Evaluation Metrics.} 

We evaluate lossy compression performance using the Bjøntegaard Delta Rate (BD-Rate) \citep{bjontegaard2001calculation} metric, which quantifies the average bitrate savings at a given level of distortion. A lower BD-Rate indicates better compression efficiency. We measure the reconstruction distortion using Peak Signal-to-Noise Ratio (PSNR) \citep{psnr}. The bitrate is assessed in bits per pixel (BPP), where uncompressed data corresponds to 24 BPP.

\paragraph{Results.} 

\begin{figure}[tb]
    \centering
    \includegraphics[width=\linewidth]{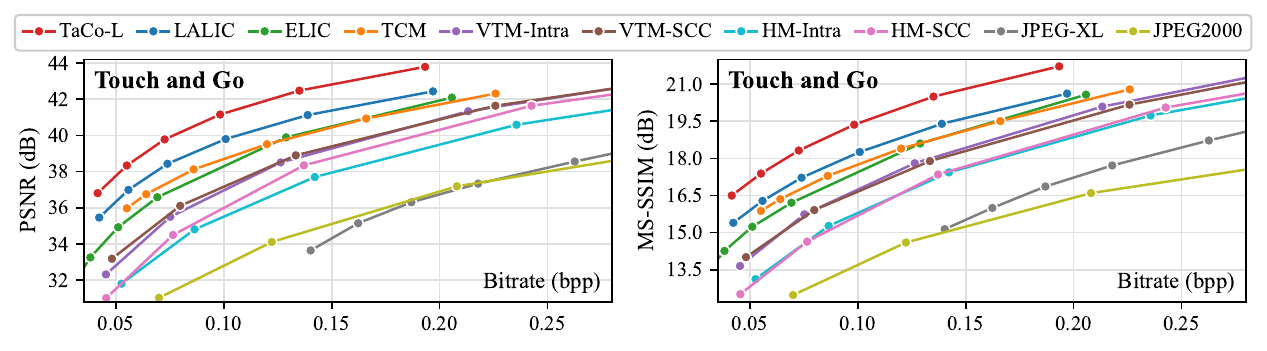}
    \vspace{-20pt}
    \caption{\footnotesize Rate-distortion curves on TouchandGo dataset, when applying intra-frame compression methods.}
    \label{fig:lossy-image-touchandgo}
\end{figure}

\cref{tab:compare-lossy-rd} benchmarks the lossy compression performance when using intra-frame compressors. Off-the-shelf intra-frame codecs like HM-Intra and VTM-Intra provide strong baselines, consistently delivering competitive performance. General-purpose codecs like JPEG2000 and JPEG-XL are included as standard baselines, but their performance is relatively poor. Neural compression methods pretrained on natural images, such as ELIC, LALIC, and TCM, show promising results on some datasets, but they fail to generalize to more structurally distinct data like ObjTac. In contrast, our TaCo-L model, trained on tactile datasets, achieves the best performance across all five datasets. It outperforms all baselines with BD-Rate reductions of -61.8\% (TouchandGo), -24.3\% (ObjectFolder), -27.4\% (YCB-Slide), and -27.0\% (ObjTac). Further, the force-based ObjTac dataset, which is derived from 3D force signals and mapped into RGB images, exhibits characteristics similar to screen content (large uniform regions and repetitive patterns). This makes screen-content-optimized codecs, VTM-SCC and HM-SCC, particularly effective on this dataset, achieving BD-Rates of -44.3\% and -44.5\%, respectively, when taking HM-Intra as the anchor. 
\cref{tab:compare-lossy} benchmarks the complexity of different lossy compression algorithms. For off-the-shelf codecs, the complexity increases along with the development of newer generations. For neural codecs, TaCo-L, adapted from the latest LALIC, achieves the best compression performance at the cost of incremental complexity, and the encoding/decoding FPS ranges from 4 FPS to 48 FPS at different resolutions. \cref{fig:lossy-image-touchandgo} presents a representative rate-distortion (RD) curve comparison on the TouchandGo dataset when using image compressors. TaCo-L achieves the SoTA performance across various bitrates. 




\begin{table}[!tb]
    \belowrulesep=0pt
    \aboverulesep=0pt
    \centering
    \renewcommand{\arraystretch}{1.1}
    \setlength{\tabcolsep}{10.5pt}
    \tabsize
    \caption{Material classification results on TouchandGo, ObjectFolder-1.0 and object classification results on YCB-Slide. Best results are in \textbf{\textcolor{darkblue}{blue bold}}, the second-best results are in bold, and the third-best in \underline{underline}.}
    \vspace{-6pt}
    \begin{tabular}{cll|cccc}
    \toprule
     & Compressor & BPP & SVM  & Random Forest  & K-NN & Linear Regression\\
    \midrule
    \multirow{5.3}{*}{\rotatebox{90}{\textbf{Touch and Go}}} 
    & \cellcolor{lightgray}{Uncompressed} 
    & \cellcolor{lightgray}{24 ($1\times$)} 
    & \cellcolor{lightgray}{\textbf{\textcolor{darkblue}{76.63\%}}} 
    & \cellcolor{lightgray}{\textbf{\textcolor{darkblue}{74.88\%}}} 
    & \cellcolor{lightgray}{\textbf{\textcolor{darkblue}{68.24\%}}}
    & \cellcolor{lightgray}{\textbf{\textcolor{darkblue}{73.51\%}}}  \\

    & VTM-Intra 
    & 0.213
    & 74.08\%
    & 72.53\% 
    & 65.06\%
    & 70.87\%\\

    & JPEG-XL 
    & 0.218
    & 70.67\%
    & 69.43\% 
    & 61.87\%
    & 67.75\%\\

    & LALIC 
    & 0.196
    & \underline{74.70\%}
    & \underline{73.07\%} 
    & \underline{65.43\%}
    & \underline{71.24\%} \\
    
    & \cellcolor{lightblue}TaCo-L (ours)
    & \cellcolor{lightblue}0.193 ($124\times$)
    & \cellcolor{lightblue} \textbf{75.12\%}
    & \cellcolor{lightblue} \textbf{73.55\%} 
    & \cellcolor{lightblue} \textbf{66.03\%}
    & \cellcolor{lightblue} \textbf{71.89\%} \\

    \midrule
    
    \multirow{5.3}{*}{\rotatebox{90}{\textbf{ObjectFolder}}} 
    & \cellcolor{lightgray}{Uncompressed} 
    & \cellcolor{lightgray}{24 ($1\times$)} 
    & \cellcolor{lightgray}{\textbf{\textcolor{darkblue}{44.11\%}}}
    & \cellcolor{lightgray}{\textbf{\textcolor{darkblue}{40.68\%}}}
    & \cellcolor{lightgray}{\textbf{\textcolor{darkblue}{37.14\%}}}
    & \cellcolor{lightgray}{\textbf{\textcolor{darkblue}{42.92\%}}}\\

    & VTM-Intra 
    & 0.384
    & \underline{42.23\%}
    & \underline{39.48\%} 
    & \underline{36.00\%}
    & \underline{40.71\%} \\

    & JPEG-XL 
    & 0.499
    & 40.27\%
    & 37.63\% 
    & 34.36\%
    & 38.74\%\\

    & LALIC 
    & 0.477
    & 41.00\%
    & 38.28\% 
    & 35.44\%
    & 39.91\%\\
    
    & \cellcolor{lightblue}TaCo-L (ours)
    & \cellcolor{lightblue}0.453 ($53\times$)
    & \cellcolor{lightblue} \textbf{43.08\%}
    & \cellcolor{lightblue} \textbf{39.85\%} 
    & \cellcolor{lightblue} \textbf{36.27\%}
    & \cellcolor{lightblue} \textbf{41.02\%} \\

    \midrule
    
    \multirow{5.3}{*}{\rotatebox{90}{\textbf{YCB-Slide}}} 
    & \cellcolor{lightgray}{Uncompressed} 
    & \cellcolor{lightgray}{24 ($1\times$)} 
    & \cellcolor{lightgray}{\textbf{\textcolor{darkblue}{98.75\%}}}
    & \cellcolor{lightgray}{\textbf{\textcolor{darkblue}{98.72\%}}}
    & \cellcolor{lightgray}{\textbf{\textcolor{darkblue}{98.58\%}}}
    & \cellcolor{lightgray}{\textbf{\textcolor{darkblue}{99.18\%}}}\\

    & VTM-Intra 
    & 0.118
    & \underline{97.36\%}
    & \underline{96.41\%} 
    & \underline{97.08\%}
    & \underline{97.24\%} \\

    & JPEG-XL 
    & 0.121
    & 94.08\%
    & 93.11\% 
    & 93.67\%
    & 93.97\%\\

    & LALIC 
    & 0.130
    & 95.67\%
    & 95.22\% 
    & 95.76\%
    & 96.23\%\\
    
    & \cellcolor{lightblue}TaCo-L (ours)
    & \cellcolor{lightblue}0.126 ($190\times$)
    & \cellcolor{lightblue} \textbf{98.01\%}
    & \cellcolor{lightblue} \textbf{97.35\%} 
    & \cellcolor{lightblue} \textbf{97.88\%}
    & \cellcolor{lightblue} \textbf{98.20\%} \\
    \bottomrule
    \end{tabular}
    \label{tab:classfication}
\end{table}

\subsection{Lossy Compression for Classification}

\paragraph{Evaluation Metrics.} 
We evaluate the semantic fidelity of lossy compression using two tactile understanding tasks: material classification (on TouchandGo and ObjectFolder-1.0) and object classification (on YCB-Slide). For each dataset, we use four standard classifiers, SVM \citep{svm}, Random Forest \citep{randomforest}, K-NN \citep{knn}, and Linear Regression \citep{linregression}, with a fixed 60\%/40\% train-test data split. The top-1 accuracy is used as evaluation metric. Four representative lossy codecs, VTM-Intra, JPEG-XL, LALIC, and TaCo-L, are used for comparison.


\paragraph{Results.} \cref{tab:classfication} As shown in \cref{tab:classfication}, all methods achieve classification performance close to the uncompressed data, despite substantial bitrate savings (e.g., from 24 bpp to as low as 0.118 bpp). On TouchandGo, TaCo-L achieves 75.12\% (SVM) and 71.89\% (Linear Regression), similar to 76.63\% and 73.51\% when using uncompressed data. On ObjectFolder, where the task is more challenging, the top-1 accuracy under SVM drops slightly from 44.11\% to 43.08\% after compression with TaCo-L. On YCB-Slide, TaCo-L also preserves superior classification accuracy (98.01\% and 98.20\% when using SVM and Linear Regression, respectively), while reducing the bitrate by $190\times$.

\begin{table}[!tb]
    \belowrulesep=0pt
    \aboverulesep=0pt
    \centering
    \renewcommand{\arraystretch}{1.1}
    \setlength{\tabcolsep}{8pt}
    \tabsize
    \caption{Evaluation results on the dexterous grasping. Best results are shown in \textbf{\textcolor{darkblue}{blue bold}}, the second-best results are denoted in bold, and the third-best in \underline{underline}. We also list the accuracy loss relative to the uncompressed data (8 bits/Byte) in parentheses.}
    \vspace{-6pt}
    \begin{tabular}{cll|ccccc}
    \toprule
    
     & Compressor & BPP  & Small Obj. &  Medium Obj.  & Large Obj. &  Deform. Obj.  & Avg\\
    \midrule
    \multirow{5.3}{*}{\rotatebox{90}{$\mathbf{s_{\text{lift}}}$}}
    & \cellcolor{lightgray}{Uncompressed} 
    & \cellcolor{lightgray}{24 } 
    & \cellcolor{lightgray}{\textbf{\textcolor{darkblue}{54.7\%}}}
    & \cellcolor{lightgray}{\textbf{\textcolor{darkblue}{67.4\%}}}
    & \cellcolor{lightgray}{\textbf{\textcolor{darkblue}{69.2\%}}}
    & \cellcolor{lightgray}{\textbf{\textcolor{darkblue}{63.9\%}}}
    & \cellcolor{lightgray}{\textbf{\textcolor{darkblue}{63.8\%}}}~\tiny{(-0.0\%)}
    \\


    & JPEG-XL  
    & 0.0505 
    & 47.2\% & 58.0\%  & 59.7\% & 55.1\%  & 55.0\%~\tiny{(-8.8\%)} \\
    

    & VTM-Intra 
    & 0.0498
    & \textbf{54.1\% }
    & \textbf{66.6\% }
    & \underline{\textbf{68.4\% }}
    & \textbf{63.1\% }
    & \textbf{63.1\% }~\tiny{(-0.7\%)}  \\

    & LALIC 
    &  0.0397 
    & 51.5\%
    & 63.4\% 
    & 65.0\%
    & 60.1\%
    & 60.0\%~\tiny{(-3.8\%)}  \\
    
    & \cellcolor{lightblue}TaCo-L (ours)
    & \cellcolor{lightblue}\textbf{0.0251 } 
    & \cellcolor{lightblue}\underline{53.1\%}
    & \cellcolor{lightblue}\underline{{65.3\%}}  
    & \cellcolor{lightblue}\underline{\textbf{68.4\%}} 
    & \cellcolor{lightblue}\underline{61.9\%} 
    & \cellcolor{lightblue}\underline{62.2\%} \tiny{(-1.6\%)} 
    \\

    \midrule
    \multirow{5.3}{*}{\rotatebox{90}{$\mathbf{s_{\text{disturb}}}$}}
    & \cellcolor{lightgray}{Uncompressed} 
    & \cellcolor{lightgray}{24} 
    & \cellcolor{lightgray}{\textbf{\textcolor{darkblue}{51.8\%}}}
    & \cellcolor{lightgray}{\textbf{\textcolor{darkblue}{65.8\%}}}
    & \cellcolor{lightgray}{\textbf{\textcolor{darkblue}{67.3\%}}}
    & \cellcolor{lightgray}{\textbf{\textcolor{darkblue}{61.8\%}}}
    & \cellcolor{lightgray}{\textbf{\textcolor{darkblue}{61.7\%}}} ~\tiny{(-0.0\%)}
    \\


    & JPEG-XL  
    & 0.0505 
    & 46.4\% & 56.1\% & 57.4\% & 52.8\%  & 53.2\%~\tiny{(-8.5\%)}  \\
    

    & VTM-Intra 
    & 0.0498
    & \textbf{52.5\%} 
    & \textbf{65.0\% }
    & \textbf{66.7\% }
    & \textbf{61.1\% } 
    & \textbf{61.3\%}~\tiny{(-0.4\%)} \\

    & LALIC 
    &  0.0397 
    & 49.9\%
    & 61.8\% 
    & 63.1\%
    & 58.0\%
    & 58.2\%~\tiny{(-3.5\%)} \\
    
    & \cellcolor{lightblue}TaCo-L (ours)
    & \cellcolor{lightblue}\textbf{0.0251}
    & \cellcolor{lightblue}\underline{51.3\%}
    & \cellcolor{lightblue}\underline{63.6\%} 
    & \cellcolor{lightblue}\underline{65.0\%}
    & \cellcolor{lightblue}\underline{59.7\%} 
    & \cellcolor{lightblue}\underline{59.9\%}~\tiny{(-1.8\%)} \\

    \bottomrule
    \end{tabular}
    \label{tab:grasp-sr}
\end{table}


\subsection{Lossy Compression for Dexterous Grasping}

\paragraph{Task Definition.}
Many contact-rich manipulation algorithms for dexterous hands rely heavily on high-fidelity tactile signals, which motivates us to conduct dexterous grasping experiment. In real-world deployment scenarios, however, tactile data compression may affect the downstream performance of such algorithms. Therefore, we introduce this experiment to evaluate the impact of tactile compression quality on a realistic, task-driven benchmark.
The goal of this task is to reach for an object, grasp and lift it. We build the simulation using Nvidia IssacSim \cite{makoviychuk2021isaac}, and use a simple DexHand13 module~\cite{paxinidexhand} equipped with eleven tactile sensors. We modify the input tactile signals by compressing them first and then feed it into a tactile-aware reinforcement learning algorithm. In total we use 100 objects to evaluate the grasping performance, consisting of 29 small objects, 41 medium objects and 22 large objects, 8 deformable objects. For the following section, we list the performance for each category.

\paragraph{Evaluation Metrics.}


To ensure robust interference capabilities during grasping, we evaluate the performance using two evaluation metrics in the simulation. One is the success rate of lifting $s_{\text{lift}}$, recorded when objects maintain stability lifted to the height of 0.1m. The other is the success rate with disturbance resistance $s_{\text{disturb}}$, measured by applying 2.5N external forces along six axes for 2 s after lifting and the object moves below 0.02 m.

\paragraph{Results.}

\cref{tab:grasp-sr} benchmarks the grasping performance across different objects. Since the tactile signal simulated by Isaac Sim is relatively sparse, the achieved compression ratio is higher (up to 1000$\times$) than what is typically attainable in the physical world (i.e. physical data achieve at most 22$\times$ (ObjTac) compression ratio). All compression methods successfully reduced the raw 24 bpp tactile signal to substantially smaller sizes, ranging from 0.025 bpp to 0.5 bpp, with only a moderate decrement in grasping success rate. Among them, TaCo-L outperforms JPEG-XL and LALIC, achieving a higher compression ratio while maintaining competitive lifting success rate of 62.2\% compared to the baseline 63.8\% and disturb-resistant grasping success rate of 59.9\% compared to the baseline 61.7\%. 
While our method TaCo-L is only approximately 1\% of VTM's performance (62.2\% vs 63.1\% and 59.9\% vs 61.3\%) in terms of task success rate, it achieves significantly higher compression efficiency by operating at nearly half the bitrate (0.0251bpp vs. 0.0498 bpp).
\section{Conclusion}

This paper introduced the \textbf{TaCo} benchmark, the first comprehensive framework for evaluating tactile data codecs. This is a suite of 30 codecs, 5 datasets and 4 types of tasks to advance the research on tactile sensing and tactile data compression. Further, we presented \textbf{TaCo-LL} and \textbf{TaCo-L}, data-driven codecs that learn the latent distribution of tactile data end-to-end. Extensive experiments demonstrate that our proposed models establish a new state-of-the-art result, outperforming existing methods across lossless/lossy compression, classification, and grasping tasks. Our work provides a critical foundation and a baseline for future research in efficient tactile perception and transmission.

\section*{ACKNOWLEDGMENT}

This work was partly supported by the NSFC (62431015, 62571317, 62501387), the Fundamental Research Funds for the Central Universities, Shanghai Key Laboratory of Digital Media Processing and Transmission under Grant 22DZ2229005, 111 project BP0719010.





{\footnotesize
\bibliography{reference}
\bibliographystyle{iclr2025_conference}
}

\newpage

\appendix
\section{Appendix}

\subsection{Baseline and Implementation Details}


We benchmark the performance of tactile compression on five representative tactile datasets: Touch and Go\footnote{\url{https://touch-and-go.github.io/}}, ObjectFolder 1.0\footnote{\url{https://objectfolder.stanford.edu/}}, SSVTP\footnote{\url{https://sites.google.com/berkeley.edu/ssvtp}}, YCB-Slide\footnote{\url{https://github.com/rpl-cmu/YCB-Slide}}, and ObjTac\footnote{\url{https://readerek.github.io/Objtac.github.io/}}, as detailed in \cref{sec:datasets}. Specifically, 70\% of the samples from Touch and Go and ObjectFolder are used for training, while the remaining 30\%, along with the full SSVTP and ObjTac datasets, are used for evaluation. 
For the Touch and Go dataset, while the official guideline recommends splitting by collection trajectories, it does not specify an exact train/test ratio or content. We followed this recommendation by grouping data at the trajectory level and applied a common 70\% and 30\% split for training and testing. Each trajectory was then decomposed into individual frames, ensuring that all frames from the same trajectory are contained entirely within either the training or the testing set, avoiding any data leakage.



%

For \textbf{TaCo-LL}, adapted from DualComp-I \citep{dualcomp}, we train it using the FusedAdam optimizer \citep{fusedadam} with a cosine annealing learning rate schedule. \citep{scheduler}, starting from $1\times10^{-4}$ and decaying to $2\times10^{-5}$ over 20 epochs. The batch size is set to 64. The model is trained with a standard cross-entropy loss:
\begin{equation}
    \mathcal{L}_{\text{TaCo-LL}}=-\sum q \log p
\end{equation}
where $q$ and $p$ are the target and predicted distributions, respectively.


For \textbf{TaCo-L}, We train our models using the Adam optimizer \citep{adam} with a batch size of 8. The model is optimized with a rate-distortion loss:
\begin{equation}
    \mathcal{L}_{\text{TaCo-L}}=R+\lambda\cdot \text{MSE}
\end{equation}
where $R$ denotes the estimated bitrate and $\lambda$ controls the trade-off between rate and distortion. For MSE-optimized models, $\lambda$ is set to $\{0.0018, 0.0067, 0.025, 0.0483\}$ to achieve different bitrates. The learning rate is set to $1\times10^{-4}$ for 40 epochs, and then decayed to $1\times10^{-5}$ for another 4 epochs. During training, input tactile images are randomly cropped or padded to $256\times256$ resolution.

These two models are trained using two NVIDIA A100 GPUs with mixed precision enabled.

\subsection{Cross-Dataset Compression Performance}

Furthermore, the above training datasets are mainly collected on rigid and lambertian objects, and we also validate compression performance on two new test datasets: Active Cloth~\citep{activeclothing} covering soft and textured objects, and ObjectFolder-2.0 comprising a wide variety of everyday 3D objects, as shown in \cref{tab:datasets}. Due to the large scale of both datasets, we conduct quick validation on approximately the first 10\% of the data from each: 10 objects from Active Cloth and 100 objects from ObjectFolder-2.0. Lossless and lossy compression are performed, as summarized in Table~\ref{tab:compare-lossless-ob2-activecloth} and Table~\ref{tab:compare-lossy-ob2-activecloth}.

\begin{table}[H]
    \centering
    \tabsize
    \belowrulesep=0pt
    \aboverulesep=0pt
    \renewcommand{\arraystretch}{1.0}
    \setlength{\tabcolsep}{8pt}
    \begin{tabular}{l|llll} 
    \toprule
    \textbf{Dataset} 
    & \textbf{\#Objects} 
    & \textbf{\#Frames} 
    & \textbf{Resolution}
    & \textbf{Sensor} 
    \\
    \midrule
    ActiveCloth~\citep{activeclothing} 
    & 153 & 494655 & $640\times480\times30\text{Hz}$ 
    & GelSight \citep{yuan2017gelsight}  \\
    ObjectFolder 2.0~\citep{gao2022ObjectFolderV2} 
    & 1000 & 76000 & $120\times160\times30\text{Hz}$
    & GelSight \citep{yuan2017gelsight}  \\
    
    \bottomrule
    \end{tabular}
    \caption{\footnotesize Introduction of two additional tactile datasets, where ActiveCloth~\citep{activeclothing} consists of 153 varied pieces of clothes and ObjectFolder-2.0~\citep{gao2022ObjectFolderV2} mainly extends ObjectFolder-1.0~\citep{gao2021objectfolder} with 100 virtualized objects to 1000 common household real objects.}
    \label{tab:datasets}
\end{table}

When comparing Active Cloth and Touch and Go at the same resolution of $640\times480$, our TaCo-LL model, with 96M parameters, achieves the best performance on both datasets. It achieves 0.723 bit/Byte (in Table~\ref{tab:compare-lossless-ob2-activecloth}) and 0.447 bit/Byte (in \cref{tab:compare-lossless-rd}), corresponding to compression ratios of $11\times$ and $18\times$, respectively. The results also suggest that soft objects in Active Cloth are more difficult to compress than rigid objects, as deformable surfaces tend to generate more complex tactile data. When comparing ObjectFolder-1.0 and ObjectFolder-2.0 at the same resolution of $120\times160$, all the compression methods basically achieve consistent results and our TaCo-LL also achieve the best performance with 2.855 bit/Byte, corresponding to compression ratios of $2.8\times$. 

These findings are further supported by the BD-Rate comparisons in Table~\ref{tab:compare-lossy-ob2-activecloth}, where TaCo-L consistently achieves the lowest BD-Rate across both ActiveCloth and ObjectFolder-2.0 datasets, outperforming state-of-the-art neural compressors such as ELIC, LALIC, and TCM.

\begin{table*}[tb]
    \belowrulesep=0pt
    \aboverulesep=0pt
    \centering
    \caption{\footnotesize { Comparison of lossless compression performance (bits/Byte) on two additional tactile datasets to validate the cross-dataset performance. The best results are highlighted in \textbf{\color{darkblue}bold blue}, second-best in \textbf{bold}, and third in \underline{underline}. For TaCo-LL, 12M/48M/96M denotes the model parameter. To show the compression performance more clearly, we also list the compression ratios relative to the uncompressed data (8 bits/Byte) in parentheses only for top three results.}}
    \tabsize
    \renewcommand{\arraystretch}{1.0}
    \setlength{\tabcolsep}{6.5pt}
    \newcommand{\centerdash}[1]{\ifx#1-\multicolumn{1}{c}{-}\else#1\fi}
    \vspace{-6pt}
    \begin{tabular}{p{0.24cm}p{4cm}|ll}
    \toprule
     & \multirow{2}{*}{\textbf{Compressor}} 
     & \multicolumn{2}{c}{\textbf{bits/Byte$\downarrow$}}\\[1pt]
     \cmidrule{3-4}

     & & ActiveCloth
     & ObjectFolder-2.0 \\
    \midrule

    & \cellcolor{lightgray}{uncompressed}        
    & \cellcolor{lightgray}{8 ($1\times$)}
    & \cellcolor{lightgray}{8 ($1\times$)}\\
    \midrule
    \multirow{9.3}{*}{\rotatebox{90}{\textbf{Off-the-Shelf}}} 
    & gzip \citep{gzip}   & 2.762 & 4.040 \\
    & zstd \citep{zstd}   & 2.771 & 4.037 \\
    & bzip2 \citep{bzip2} & 2.771 & 4.103 \\
    & FLIF \citep{flif}   & 0.882 & 3.831 \\
    & BPG  \citep{bpg}    & 1.645 & 3.792 \\
    & WebP \citep{webp}   & 1.063 & 3.676 \\
    & JPEG-XL \citep{jpegxl}    & \underline{0.841} ($9.5\times$) & 3.659 \\
    & JPEG2000 \citep{jpeg2000} & 1.804 & 4.061 \\
    & PNG \citep{png}    & 2.667 & 4.035 \\
    \midrule

    \multirow{8.3}{*}{\rotatebox{90}{\textbf{Neural}}}
    & DLPR \citep{dlpr} & 1.453 & 3.852\\
    & P2LLM \citep{p2llm}  & 2.193 & 3.470 \\
    & Llama3* \citep{lmic} & 2.620 & 3.659 \\
    & RWKV* \citep{lmic}   & 2.640 & 3.800 \\
    & DualComp-I \citep{dualcomp} & 1.158 & 3.308\\

    & \cellcolor{lightpink}TaCo-LL-12M (ours)
    & \cellcolor{lightpink}1.059
    & \cellcolor{lightpink}\underline{3.179} ($2.5\times$) \\
    
    & \cellcolor{lightpink}TaCo-LL-48M (ours) 
    & \cellcolor{lightpink}\textbf{0.816} ($10\times$)
    & \cellcolor{lightpink}\textbf{3.002}  ($2.7\times$) \\
    
    & \cellcolor{lightpink}TaCo-LL-96M (ours) 
    & \cellcolor{lightpink}\textbf{\color{darkblue}0.723} ($11\times$)
    & \cellcolor{lightpink}\textbf{\color{darkblue}2.855} ($2.8\times$) \\
    \bottomrule    
    \end{tabular}
    \label{tab:compare-lossless-ob2-activecloth}
\end{table*}

\begin{table*}[tb]
    \belowrulesep=0pt
    \aboverulesep=0pt
    \centering
    \caption{\footnotesize Lossy compression performance (BD-Rate) on two additional tactile datasets to validate the cross-dataset performance. The best results are shown in \textbf{\textcolor{darkblue}{blue bold}}, the second-best in \textbf{bold}, and the third-best in \underline{underline}. For TaCo, 12M/48M/96M denotes the model parameter.}
    \tabsize
    \renewcommand{\arraystretch}{1.0}
    \setlength{\tabcolsep}{8.5pt}
    \newcommand{\centerdash}[1]{\ifx#1-\multicolumn{1}{c}{-}\else#1\fi}
    \vspace{-6pt}
    \begin{tabular}{p{0.3cm}p{3.8cm}|ll}
    \toprule
     & \multirow{2}{*}{\textbf{Compressor}} 
     & \multicolumn{2}{c}{\textbf{BD-Rate (\%)}$\downarrow$}\\[1pt]
     \cmidrule{3-4}

     & 
     & ActiveCloth & ObjectFolder-2.0 \\
    \midrule

    \multirow{6.3}{*}{\rotatebox{90}{\textbf{Off-the-Shelf}}} 
    & HM-Intra \citep{hevc}  & 0\%      & 0\% \\
    & HM-SCC \citep{hevc}    & -12.9\%  & 2.2\% \\
    & VTM-Intra \citep{vvc}  & -28.0\%  & \textbf{-21.0\%} \\
    & VTM-SCC \citep{vvc}    & -26.1\%  & \underline{-19.3\%} \\
    & JPEG-XL \citep{jpegxl} & 46.9\%   & 80.7\% \\
    & JPEG2000 \citep{jpeg2000} & 86.5\% & 79.0\% \\
    \midrule

    \multirow{4.3}{*}{\rotatebox{90}{\textbf{Neural}}}
    & ELIC \citep{elic}   & \textbf{-57.1\%} & 3.2\% \\
    & LALIC \citep{lalic} & \underline{-54.8\%}  & 2.8\%  \\
    & TCM \citep{tcm}     & -49.8\% & 23.7\% \\
    
    & \cellcolor{lightblue}TaCo-L (Ours)
    & \cellcolor{lightblue}\textbf{\textcolor{darkblue}{-65.4\%}}
    & \cellcolor{lightblue}\textbf{\textcolor{darkblue}{-26.4\%}} \\

    \bottomrule    
    \end{tabular}
    \label{tab:compare-lossy-ob2-activecloth}
\end{table*}


\subsection{Cross-object Compression Performance}


\begin{table*}[!tb]
    \belowrulesep=0pt
    \aboverulesep=0pt
    \centering
    \caption{\footnotesize {Cross-object lossless compression performance (bits/Byte) on RIGID objects. The best results are highlighted in \textbf{\color{darkblue}bold blue}, the second-best in \textbf{bold}, and the third in \underline{underline}. In TouchandGo, Tree and Wood are training objects, while Concrete is unseen. The three objects in ObjTac are all unseen objects ($^*$).}}
    \tabsize
    \renewcommand{\arraystretch}{1.0}
    \setlength{\tabcolsep}{9.3pt}
    \newcommand{\centerdash}[1]{\ifx#1-\multicolumn{1}{c}{-}\else#1\fi}
    \vspace{-6pt}
    \begin{tabular}{p{0.2cm}p{3.5cm}|lll | lll}
    \toprule
     & \multirow{2}{*}{\textbf{Compressor}} 
     & \multicolumn{5}{c}{\textbf{bits/Byte$\downarrow$}}\\[1pt]
     \cmidrule{3-8}
     &
     & \multicolumn{3}{c|}{\textbf{Touch and Go}} 
     & \multicolumn{3}{c}{\textbf{ObjTac}}  \\
     
     & 
     & Tree
     & Wood
     & Concrete$^*$ 
     & Stone$^*$ 
     & Pebble$^*$ 
     & Tile$^*$  \\
    \midrule

    & \cellcolor{lightgray}{uncompressed}        
    & \cellcolor{lightgray}{8 ($1\times$)}
    & \cellcolor{lightgray}{8 ($1\times$)} 
    & \cellcolor{lightgray}{8 ($1\times$)} 
    & \cellcolor{lightgray}{8 ($1\times$)} 
    & \cellcolor{lightgray}{8 ($1\times$)}
    & \cellcolor{lightgray}{8 ($1\times$)} \\

    \cmidrule{1-8}
    \multirow{9.3}{*}{\rotatebox{90}{\textbf{Off-the-Shelf}}} 
    & gzip \citep{gzip}   
    & 2.531 & 2.082 & 2.214 & 1.100 & 0.943 & 0.529 \\
    & zstd \citep{zstd}   
    & 2.327 & 2.033 & 2.080 & 1.098 & 0.930 & 0.505 \\
    & bzip2 \citep{bzip2} 
    & 2.486 & 2.068 & 2.186 & 1.123 & 0.976 & 0.541 \\
    & FLIF \citep{flif}   
    & 0.865 & 0.737 & \underline{0.782} & \underline{0.697} & \underline{0.648} & \underline{0.303} \\
    & BPG  \citep{bpg}    
    & 1.395 & 1.141 & 1.246 & 1.061 & 0.847 & 0.453 \\
    & WebP \citep{webp}   
    & 1.000 & 0.855 & 0.924 & 0.848 & 0.720 & 0.387 \\
    & JPEG-XL \citep{jpegxl}    
    & \underline{0.796} & \textbf{0.670} & \textbf{0.730} & 0.742 & 0.656 & 0.372 \\
    & JPEG2000 \citep{jpeg2000} 
    & 1.617 & 1.421 & 1.540 & 1.181 & 0.990 & 0.500 \\
    & PNG \citep{png}    
    & 2.765 & 2.249 & 2.390 & 1.097 & 0.939 & 0.527  \\
    \cmidrule{1-8}

    \multirow{8.3}{*}{\rotatebox{90}{\textbf{Neural}}}
    & DLPR \citep{dlpr} 
    & 1.127 & 0.935 & 1.062 & 0.906 & 0.917 & 0.551 \\
    & P2LLM \citep{p2llm} 
    & 1.946 & 1.475 & 1.832 & 0.933 & 0.911 & 0.534 \\
    & Llama3* \citep{lmic} 
    & 2.479 & 2.010 & 2.145 & 1.098 & 1.102 & 0.809 \\
    & RWKV* \citep{lmic}   
    & 2.558 & 2.120 & 2.396 & 1.175 & 1.110 & 0.832 \\
    & DualComp-I \citep{dualcomp} 
    & 0.840 & 0.726 & 0.857 & 0.810 & 0.685 & 0.339 \\
    \cmidrule{2-8}

    & \cellcolor{lightpink}TaCo-LL-12M (ours)
    & \cellcolor{lightpink}0.810
    & \cellcolor{lightpink}0.704
    & \cellcolor{lightpink}0.815
    & \cellcolor{lightpink}0.796
    & \cellcolor{lightpink}0.680
    & \cellcolor{lightpink}0.336
    \\
    
    & \cellcolor{lightpink}TaCo-LL-48M (ours) 
    & \cellcolor{lightpink}\textbf{0.719}
    & \cellcolor{lightpink}\textbf{0.611}
    & \cellcolor{lightpink}\textbf{0.730}
    & \cellcolor{lightpink}\textbf{0.635}
    & \cellcolor{lightpink}\textbf{0.627}
    & \cellcolor{lightpink}\textbf{0.300}
    \\
    
    & \cellcolor{lightpink}TaCo-LL-96M (ours) 
    & \cellcolor{lightpink}\textbf{\color{darkblue}0.607 } 
    & \cellcolor{lightpink}\textbf{\color{darkblue}0.598 } 
    & \cellcolor{lightpink}\textbf{\color{darkblue}0.700 }
    & \cellcolor{lightpink}\textbf{\color{darkblue}0.590}
    & \cellcolor{lightpink}\textbf{\color{darkblue}0.596 }
    & \cellcolor{lightpink}\textbf{\color{darkblue}0.288}
    \\
    \bottomrule    
    \end{tabular}
    \label{tab:compare-lossless-cross-obj-rigid}
\end{table*}

\begin{table*}[!tb]
    \belowrulesep=0pt
    \aboverulesep=0pt
    \centering
    \caption{\footnotesize { Cross-object lossless compression performance (bits/Byte) on SOFT objects. The best results are highlighted in \textbf{\color{darkblue}bold blue}, the second-best in \textbf{bold}, and the third in \underline{underline}. All these objects are unseen ($^*$).}}
    \tabsize
    \renewcommand{\arraystretch}{1.0}
    \setlength{\tabcolsep}{6.8pt}
    \newcommand{\centerdash}[1]{\ifx#1-\multicolumn{1}{c}{-}\else#1\fi}
    \vspace{-6pt}
    \begin{tabular}{p{0.15cm}p{3.4cm}|lll | lll}
    \toprule
     & \multirow{2}{*}{\textbf{Compressor}} 
     & \multicolumn{5}{c}{\textbf{bits/Byte$\downarrow$}}\\[1pt]
     \cmidrule{3-8}
     &
     & \multicolumn{3}{c|}{\textbf{Active Cloth}} 
     & \multicolumn{3}{c}{\textbf{ObjTac}}  \\

     &
     & Cloth-12$^*$
     & Cloth-29$^*$
     & Cloth-33$^*$  
     & Sponge$^*$ 
     & Jeans$^*$ 
     & Leather Bag$^*$  \\

    \midrule

    & \cellcolor{lightgray}{uncompressed}        
    & \cellcolor{lightgray}{8 ($1\times$)}
    & \cellcolor{lightgray}{8 ($1\times$)} 
    & \cellcolor{lightgray}{8 ($1\times$)} 
    & \cellcolor{lightgray}{8 ($1\times$)} 
    & \cellcolor{lightgray}{8 ($1\times$)}
    & \cellcolor{lightgray}{8 ($1\times$)} \\

    \cmidrule{1-8}
    \multirow{9.3}{*}{\rotatebox{90}{\textbf{Off-the-Shelf}}} 
    & gzip \citep{gzip}   
    & 3.540 & 1.902 & 3.609 & 0.214 & 0.178 & 0.129 \\
    & zstd \citep{zstd}   
    & 3.550 & 1.911 & 3.619 & 0.210 & 0.173 & 0.128 \\
    & bzip2 \citep{bzip2} 
    & 3.552 & 1.907 & 3.619 & 0.252 & 0.206 & 0.144 \\
    & FLIF \citep{flif}   
    & \underline{1.097} & \textbf{0.668} & \textbf{1.101} & 0.106 & 0.079 & \textbf{0.071} \\
    & BPG  \citep{bpg}    
    & 2.043 & 1.194 & 2.064 & 0.207 & 0.148 & 0.144 \\
    & WebP \citep{webp}   
    & 1.319 & 0.802 & 1.323 & 0.178 & 0.092 & \textbf{\color{darkblue}0.065 }\\
    & JPEG-XL \citep{jpegxl}    
    & \textbf{1.055} & \textbf{\color{darkblue}0.621} & \textbf{\color{darkblue}1.057} & 0.106 & \textbf{0.076} & 0.088 \\
    & JPEG2000 \citep{jpeg2000} 
    & 2.143 & 1.419 & 2.140 & 0.246 & 0.263 & 0.204 \\
    & PNG \citep{png}    
    & 3.549 & 1.586 & 3.619 & 0.213 & 0.197 & 0.165 \\
    \cmidrule{1-8}

    \multirow{8.3}{*}{\rotatebox{90}{\textbf{Neural}}}
    & DLPR \citep{dlpr} 
    & 1.877 & 1.590 & 1.985 & 0.148 & 0.094 & 0.150 \\
    & P2LLM \citep{p2llm} 
    & 2.033 & 1.724 & 2.082 & 0.170 & 0.142 & 0.143 \\
    & Llama3* \citep{lmic} 
    & 3.147 & 1.883 & 3.251 & 0.492 & 0.185 & 0.158 \\
    & RWKV* \citep{lmic}   
    & 3.219 & 1.890 & 3.238 & 0.510 & 0.179 & 0.163 \\
    & DualComp-I \citep{dualcomp} 
    & 1.696 & 1.125 & 2.000 & \underline{0.105} & 0.109 & 0.110 \\
    \cmidrule{2-8}
    
    & \cellcolor{lightpink}TaCo-LL-12M (ours)  
    & \cellcolor{lightpink}1.710  
    & \cellcolor{lightpink}1.147  
    & \cellcolor{lightpink}1.991
    & \cellcolor{lightpink}0.106
    & \cellcolor{lightpink}0.100
    & \cellcolor{lightpink}0.113
    \\

    & \cellcolor{lightpink}TaCo-LL-48M (ours)
    & \cellcolor{lightpink}1.332
    & \cellcolor{lightpink}0.877
    & \cellcolor{lightpink}1.930
    & \cellcolor{lightpink}\textbf{0.100}
    & \cellcolor{lightpink}\underline{0.078}
    & \cellcolor{lightpink}0.095
    \\

    & \cellcolor{lightpink}TaCo-LL-96M (ours) 
    & \cellcolor{lightpink}\textbf{\color{darkblue}1.016}
    & \cellcolor{lightpink}\underline{0.725}
    & \cellcolor{lightpink}\underline{1.255}
    & \cellcolor{lightpink}\textbf{\color{darkblue}0.097}
    & \cellcolor{lightpink}\textbf{\color{darkblue}0.053}
    & \cellcolor{lightpink}\underline{0.079}
    \\
    \bottomrule    
    \end{tabular}
    \label{tab:compare-lossless-cross-obj-soft}
\end{table*}

\begin{figure}[h]
    \centering
    \includegraphics[width=0.9\linewidth]{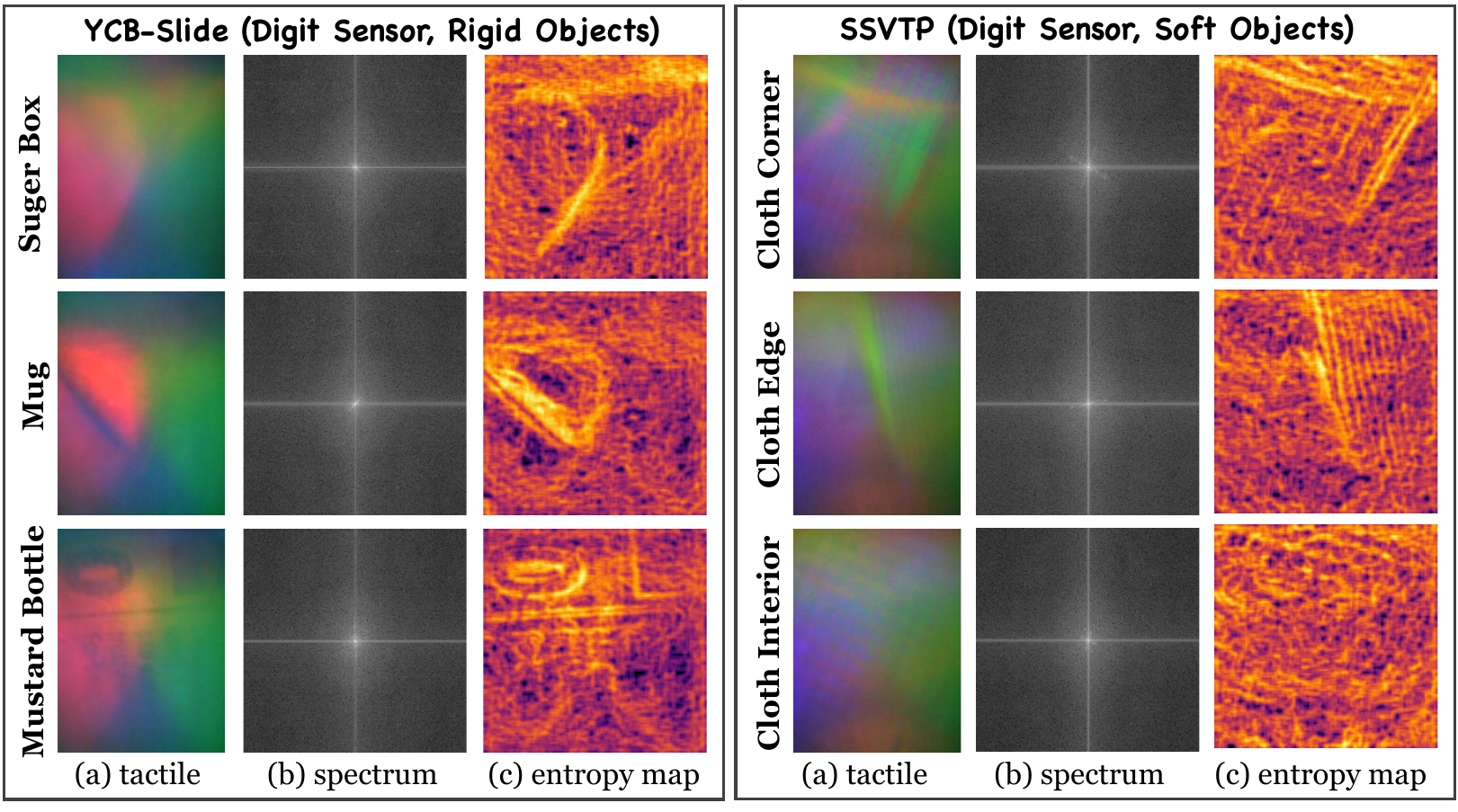}\\
    \includegraphics[width=0.9\linewidth]{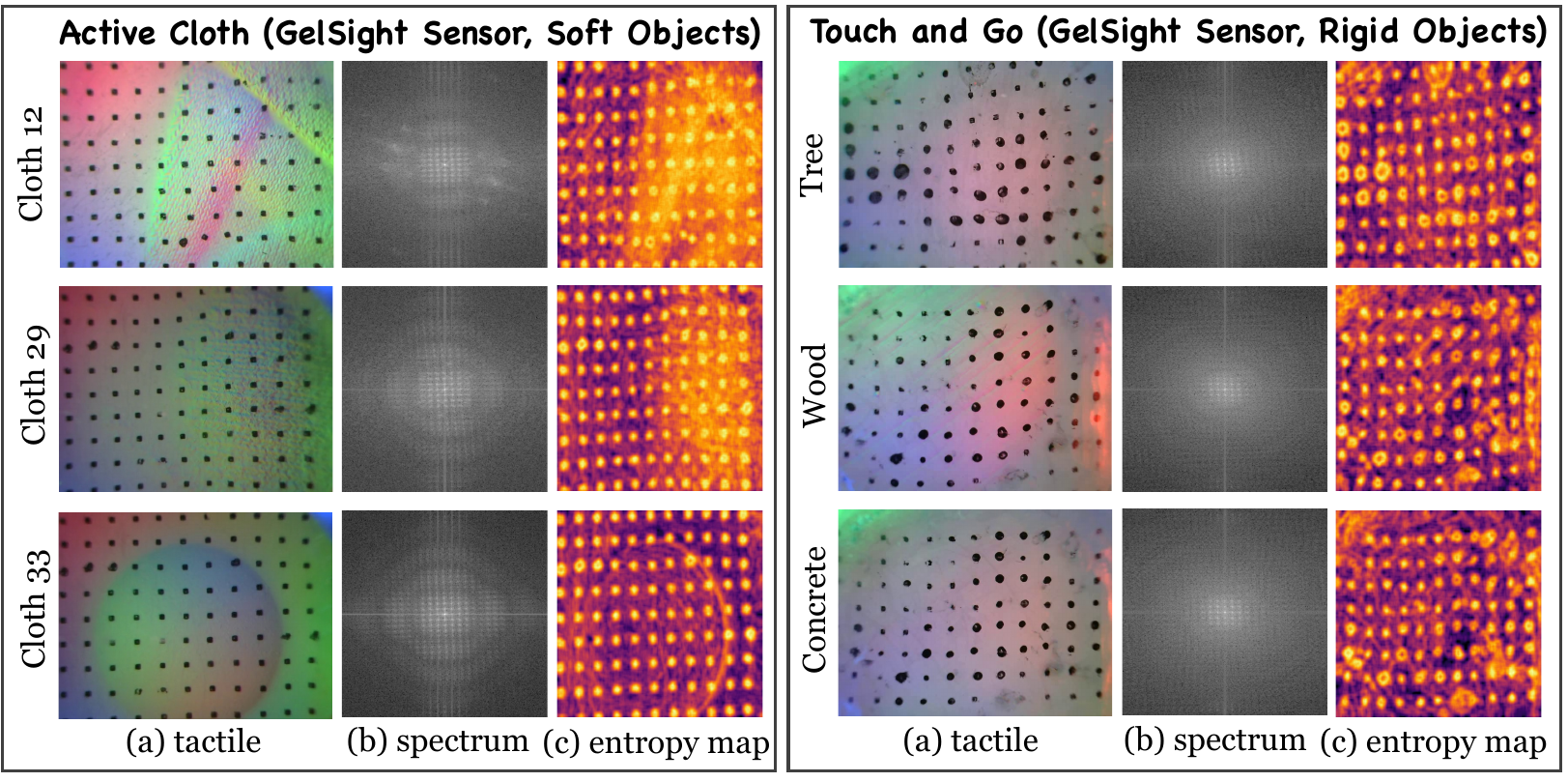}\\
    \includegraphics[width=0.9\linewidth]{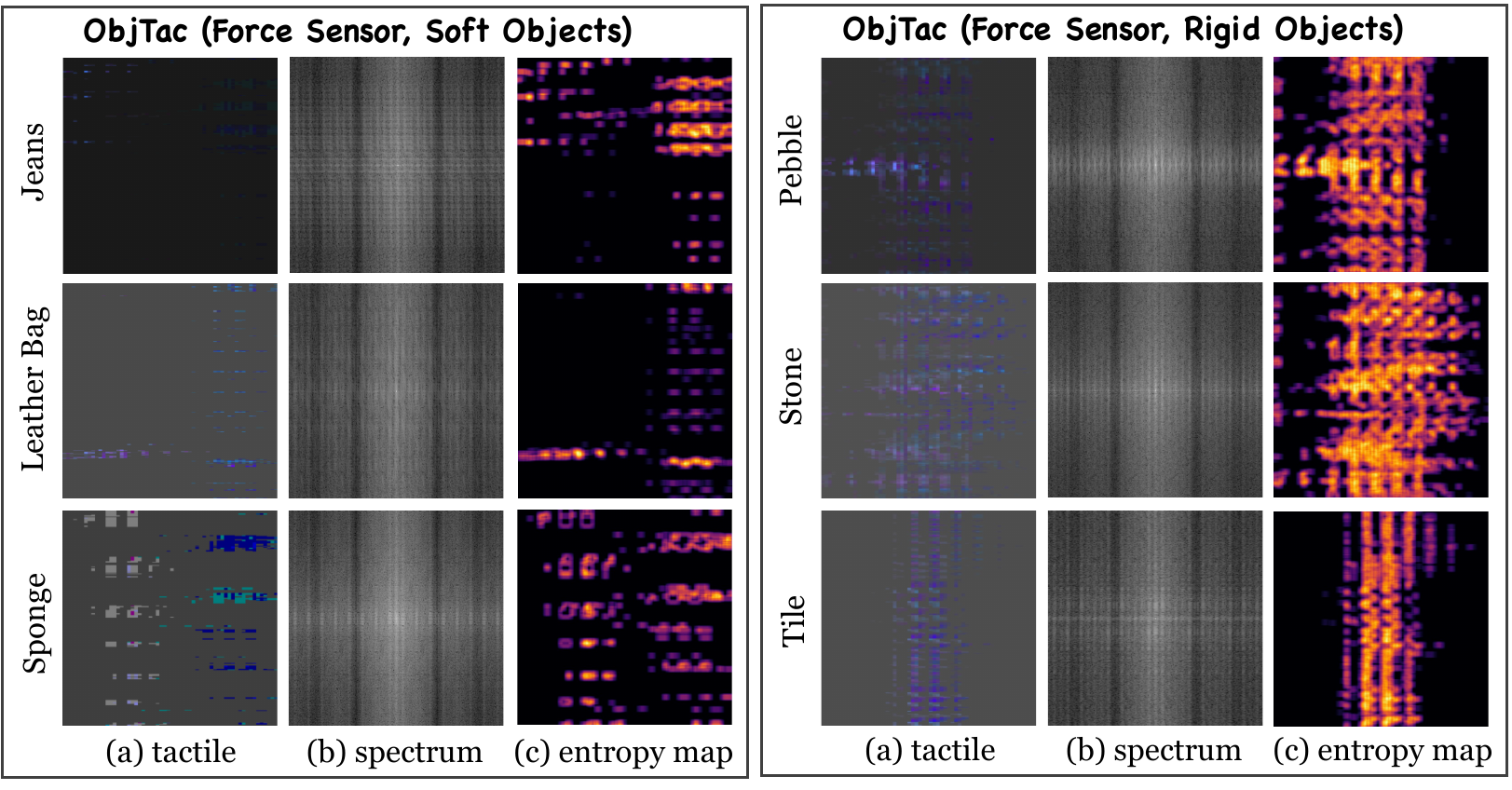}
    \caption{Visualization of tactile data characteristics across different datasets, sensors, and object types. Each subfigure displays the raw tactile image, its frequency spectrum, and the corresponding entropy map.
    }
    \label{fig:visual-cross-object}
\end{figure}

\cref{tab:compare-lossless-cross-obj-rigid} and \cref{tab:compare-lossless-cross-obj-soft} present a evaluation of cross-object lossless compression performance (in bits/Byte). \cref{tab:compare-lossless-cross-obj-rigid} focuses on rigid objects from TouchandGo and ObjTac datasets, while \cref{tab:compare-lossless-cross-obj-soft} evaluates soft objects from ActiveCloth and ObjTac datasets. A key observation is that compression performance is influenced primarily by the type of sensor modality, as evidenced by consistent trends within datasets from the same source (e.g., ActiveCloth vs. TouchandGo). However, the physical properties of the object (like rigidity or softness) also have an obvious impact. 


\subsection{Analysis of Tactile Data Characteristic}



\cref{fig:visual-cross-object} includes samples from YCB-Slide (Digit sensor, rigid objects such as Sugar Box, Mug, and Mustard Bottle), SSVTP (Digit sensor, soft cloth objects like Cloth Corner and Interior), Active Cloth (GelSight sensor, soft fabrics including Cloth-12, 29, and 33), Touch and Go (GelSight sensor, rigid surfaces such as Tree, Wood, and Concrete), and ObjTac (Force sensor, both soft objects like Jeans, Leather Bag, Sponge, and rigid ones like Pebble, Stone, Tile). As depicted, the entropy maps and FFT spectra show that tactile images are dominated by low-frequency energy with highly repetitive, grid-like spatial structures. These signals exhibit strong directional patterns and locally predictable regions, leading to sparse residuals after prediction. As a result, block-based lossy codecs such as SCC perform especially well, since their intra prediction, palette modes, and transform coding are optimized for smooth, structured, and repetitive content. The same properties also explain the behavior of lossless codecs: low entropy regions compress extremely well, while periodic patterns favor context-based or LZ-type entropy models.


\subsection{More Lossy Compression Performance for Human Vision}

In addition to intra-frame compression methods, \cref{tab:compare-lossy-video} benchmarks the use of video codecs for compressing tactile data, focusing on their ability to exploit inter-frame redundancy. It can be seen that DCVC-RT achieves the best performance on most dataset, followed by DCVC-FM and VVenC. Since each row of an ObjTac image corresponds to a distinct timestamp, the dataset does not have video format and video compression evaluation. 

\begin{table*}[h]
    \belowrulesep=0pt
    \aboverulesep=0pt
    \centering
    \caption{\footnotesize Evaluation on lossy compression performance with regard to intra-frame and inter-frame correlations. The best results are denoted in \textbf{bold}, and the second-best in \underline{underline}}
    \tabsize
    \renewcommand{\arraystretch}{1.0}
    \setlength{\tabcolsep}{8.5pt}
    \newcommand{\centerdash}[1]{\ifx#1-\multicolumn{1}{c}{-}\else#1\fi}
    \vspace{-6pt}
    \begin{tabular}{p{0.22cm}p{4.2cm}|llll}
    \toprule
     & \multirow{2}{*}{\textbf{Compressor}} 
     & \multicolumn{4}{c}{\textbf{BD-Rate (\%)}$\downarrow$} \\ [1pt]
     \cmidrule{3-6}

     & 
     & TouchandGo
     & ObjectFolder
     & SSVTP
     & YCB-Slide\\
    \midrule

    \multirow{3.3}{*}{\rotatebox{90}{\makecell{\textbf{Off-the-}\\\textbf{Shelf}}}} 
    & x265 \citep{hevc}   & 0\%     & 0\%     & 0\%     & 0\% \\ 
    & SVT-AV1 \citep{av1} & -40.6\% & -34.2\% & -28.2\% & -12.1\% \\
    & VVenC \citep{vvc}   & -67.6\% & -16.4\% & -33.6\% & -52.2\% \\
    \cmidrule{1-6}
    \multirow{3.3}{*}{\rotatebox{90}{\textbf{Neural}}}
    & DCVC-DC \citep{dcvc-dc} & -75.6\% & -12.2
    \% & -20.4\% & -27.1\% \\
    & DCVC-FM \citep{dcvc-fm} & \textbf{-80.0\%} & \underline{-43.8\%} & \underline{-45.2\%} & \underline{-58.1\%} \\
    & DCVC-RT \citep{dcvc-rt} & \underline{-78.1\%} & \textbf{-48.8\%} & \textbf{-50.9\%} & \textbf{-65.5\%} \\

    
    \bottomrule    
    \end{tabular}
    \label{tab:compare-lossy-video}
\end{table*}

\begin{figure}[h]
    \centering
    \includegraphics[width=\linewidth]{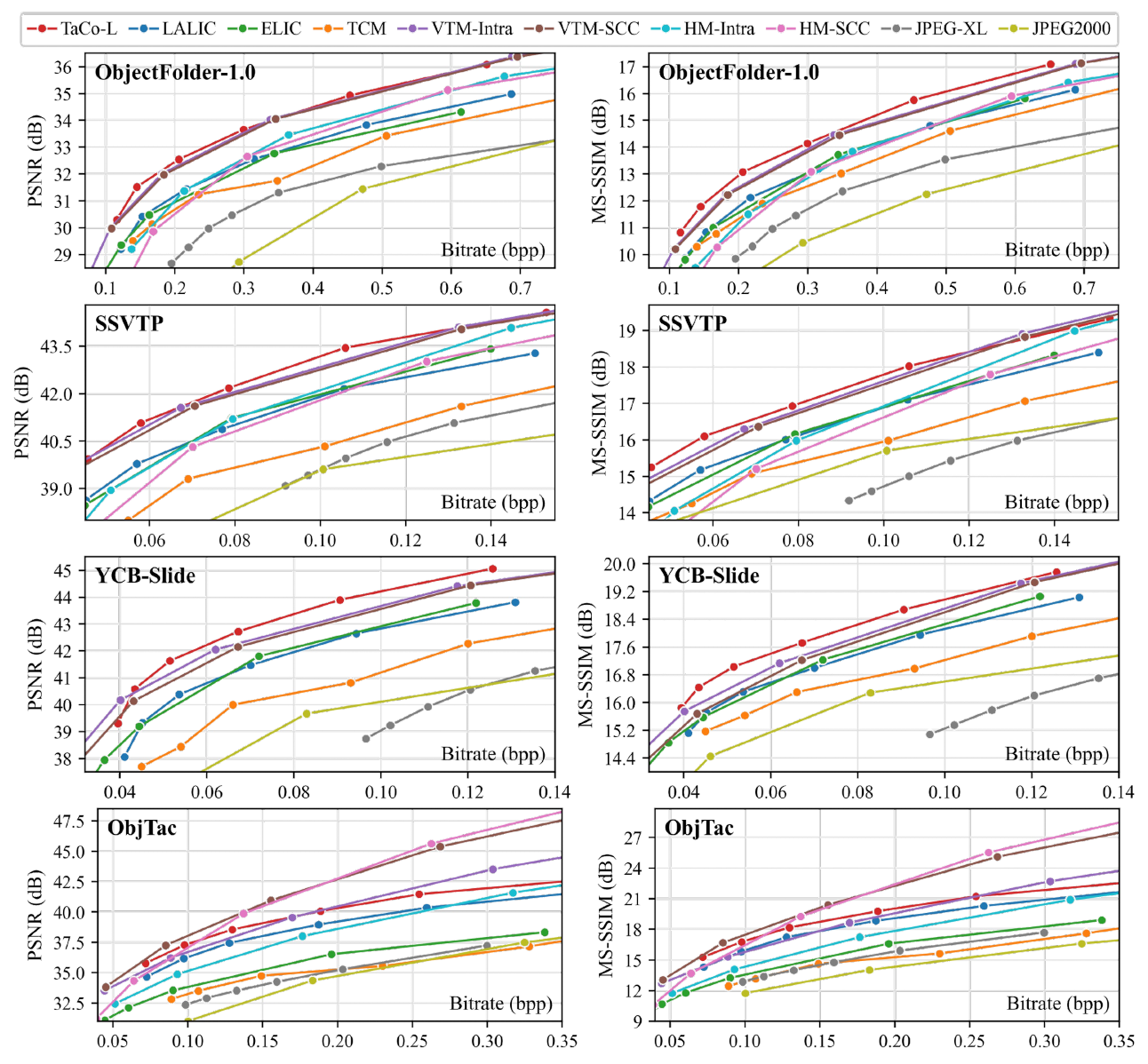}
    \caption{\footnotesize Rate-distortion performance across four tactile datasets when treating tactile data as images.}
    \label{fig:lossy-image-rd}
\end{figure}

\begin{figure}[h]
    \centering
    \includegraphics[width=\linewidth]{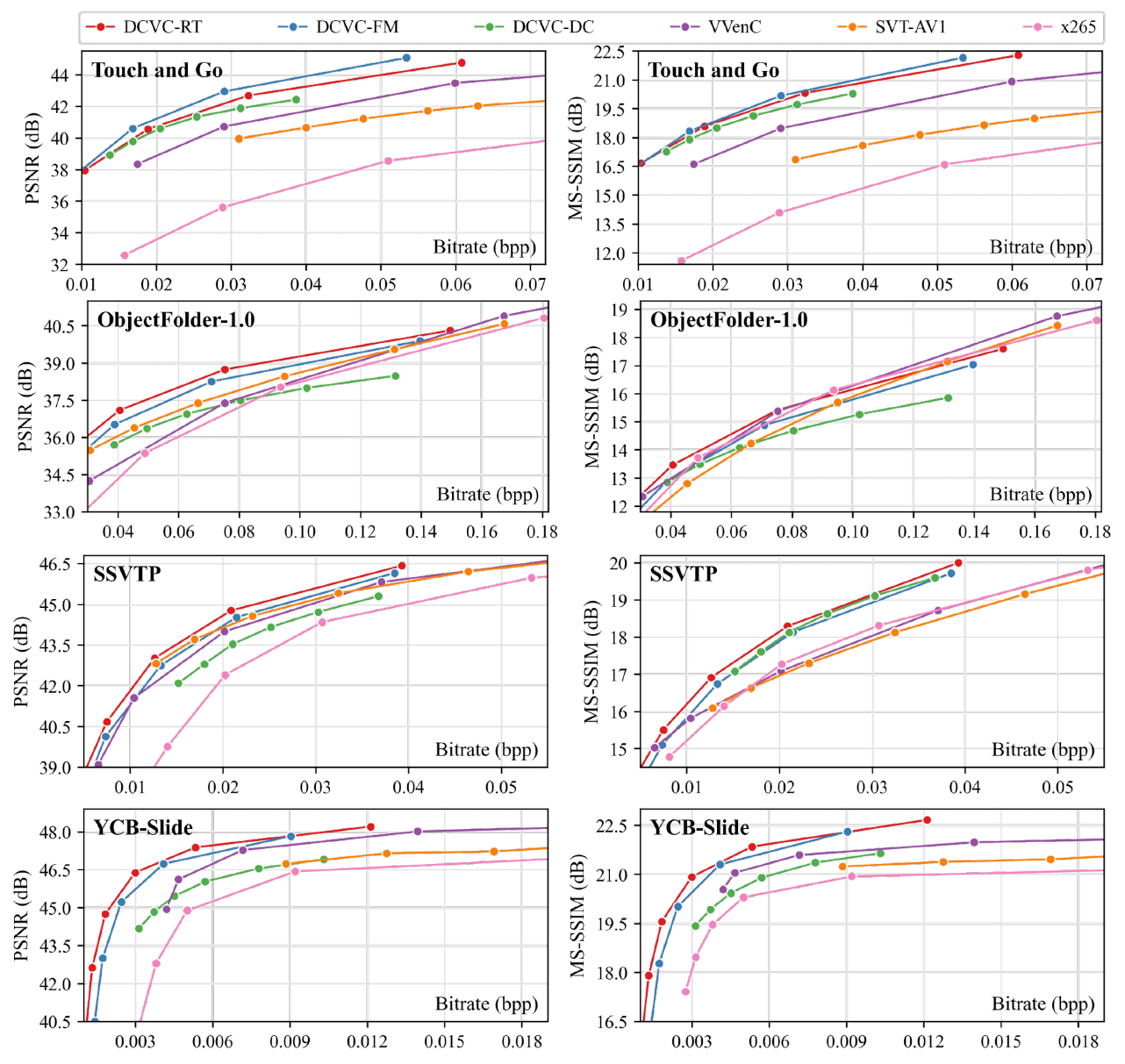}
    \caption{Rate-distortion performance across four tactile datasets when treating tactile data as videos.}
    \label{fig:lossy-video-rd}
\end{figure}

To complement the BD-Rate results in \cref{tab:compare-lossy} and \cref{tab:compare-lossy-video}, we present full rate-distortion (RD) curves for each dataset in both image-based (\cref{fig:lossy-image-rd}) and video-based (\cref{fig:lossy-video-rd}) compression settings, using PSNR and MS-SSIM as distortion metrics. For the TouchandGo dataset, image-based RD curves are shown in \cref{fig:lossy-image-touchandgo} of the main text. 


 These curves provide a more detailed view of compression performance across bitrates. In the image-based setting, JPEG-XL and JPEG2000 consistently result to relatively poor performance. TaCo-L consistently achieves the best performance across all datasets except ObjTac, where screen-content-coding (SCC) modes in VTM and HM are particularly effective due to the screen-content-like patterns. In the video-based setting, neural codecs like DCVC variants outperform traditional video codecs like x265, SVT-AV1 and VVenC, especially in the low-bitrate region.



\begin{figure}[h]
    \centering
    \includegraphics[width=\linewidth]{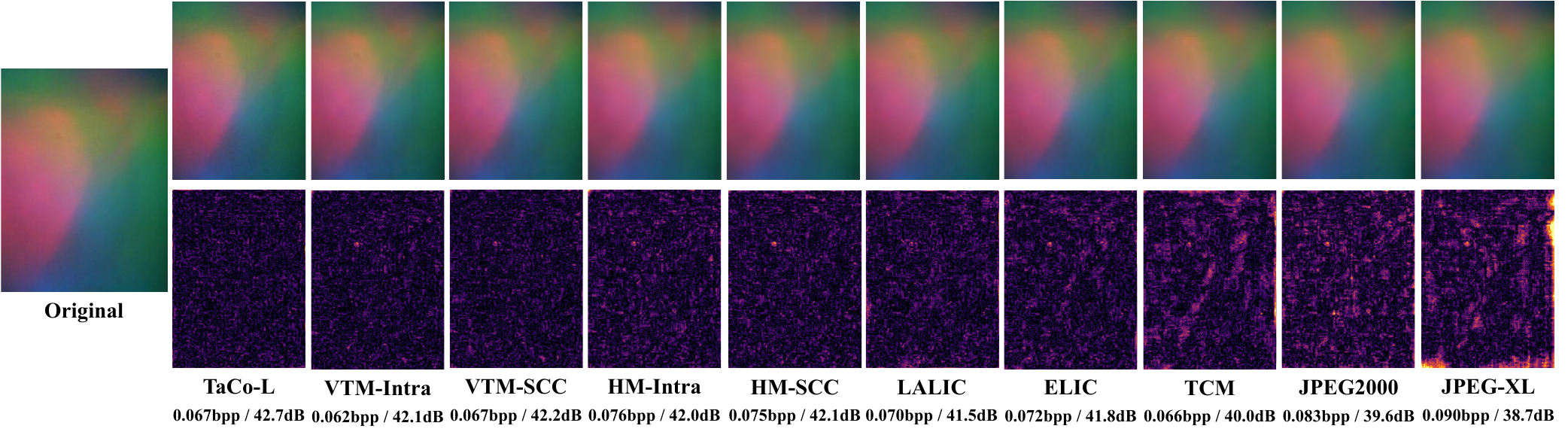}
    \caption{Visualization of reconstructed tactile images (top row) and their corresponding per-pixel root mean squared error (RMSE) maps (bottom row) on the YCB-Slide dataset. The RMSE maps highlight local reconstruction errors, with brighter regions indicating larger residuals.}
    \label{fig:visual-ycbslide}
\end{figure}

Aside from objective metrics, We also use the YCB-Slide dataset as an example and provide per-pixel RMSE error maps of the reconstructed tactile signals in \cref{fig:visual-ycbslide}. As discussed in \cref{fig:teaser} of the main paper, although tactile images carry meaningful physical information, their visual appearance is often unintuitive for human interpretation. Therefore, instead of relying on perceptual inspection, we quantify local reconstruction discrepancies using the pixel-wise RMSE,
\begin{equation}
    \text{RMSE}(x,\hat{x})=\sqrt{(x-\hat{x})^2}.
\end{equation}
As shown in \cref{fig:visual-ycbslide}, all methods produce relatively low reconstruction errors even at low bitrates (below 0.1 bpp, i.e., over $240\times$ compression), indicating that lossy compression at relatively high ratios is generally acceptable for human-viewing purposes.



\subsection{More Lossy Compression Results for Classification}


\begin{figure}[h]
    \centering
    \includegraphics[width=\linewidth]{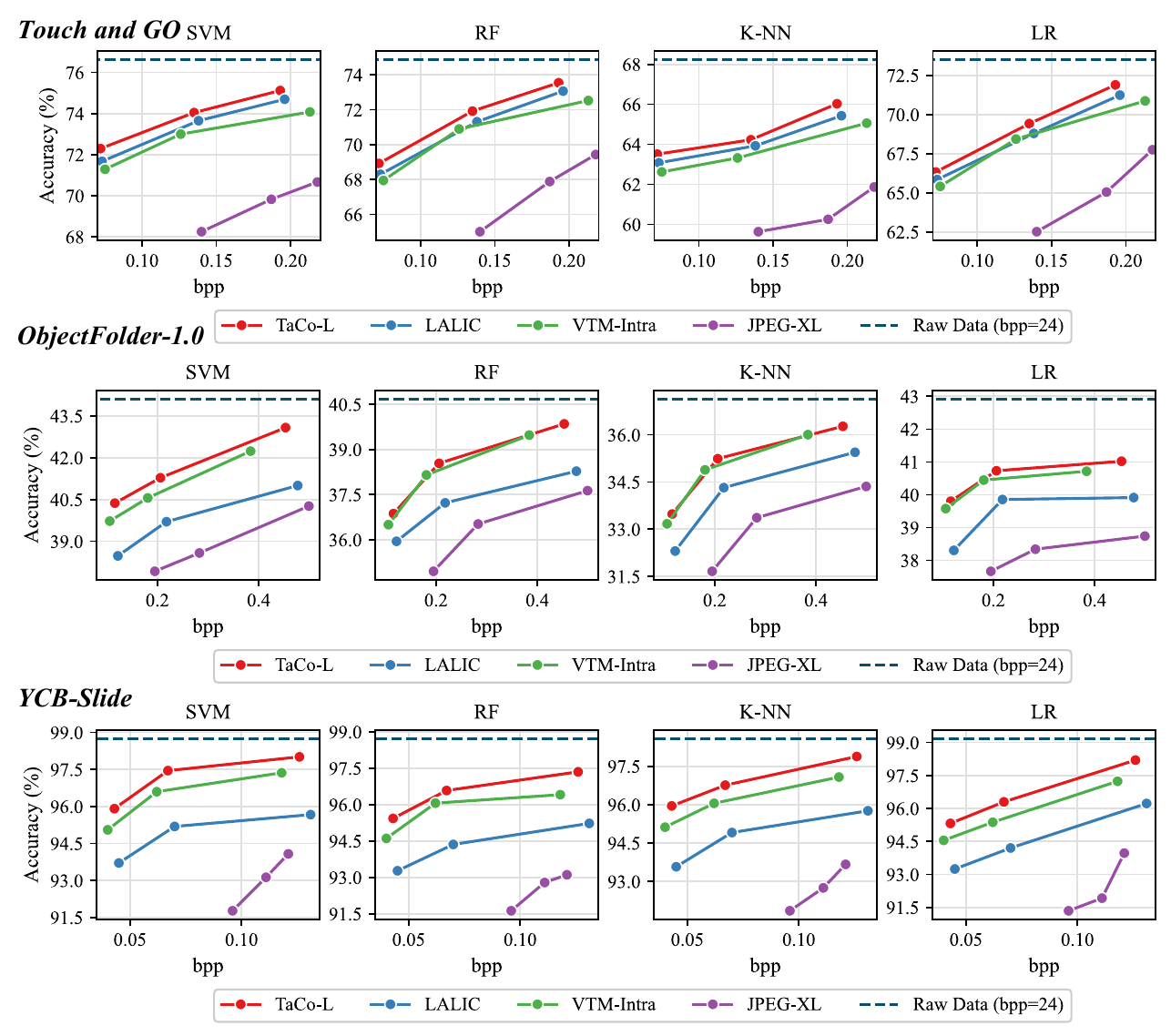}
    \caption{Bpp-accuracy curves for material classification task on the TouchandGo and ObjectFolder-1.0 datasets, and object classification task on the YCB-Slide dataset.}
    \label{fig:touchandgo-clf}
\end{figure}





To complement the results in \cref{tab:classfication} and illustrate the full-bitrate performance, we present the bitrate-accuracy curves on Touch and Go, ObjectFolder-1.0, and YCB-Slides dataset, as shown in \cref{fig:touchandgo-clf}. Each curve shows how classification accuracy changes as the bitrate varies, with dotted lines indicating the performance on uncompressed data (24 bpp). These plots provide a more comprehensive view of semantic preservation across different compression levels.

Specifically, the bitrate is varied by adjusting the quantization parameter (QP) for each compressor. For each classification task, we split the reconstructed data into 60\% for training and 40\% for testing, and apply four standard classifiers (SVM, Random Forest, K-NN, and Linear Regression) to perform material classification (TouchandGo and ObjectFolder) or object classification (YCB-Slide). Overall, even at over $200\times$ compression, the impact on classification accuracy remains minor for all methods, suggesting that lossy compression can be applied without substantially compromising downstream understanding tasks. Among them, TaCo-L consistently achieves the highest accuracy across the full bitrate range, and closely approaches the accuracy of raw data (24 bpp). 



\begin{figure}[h]
    \centering
    \includegraphics[width=\linewidth]{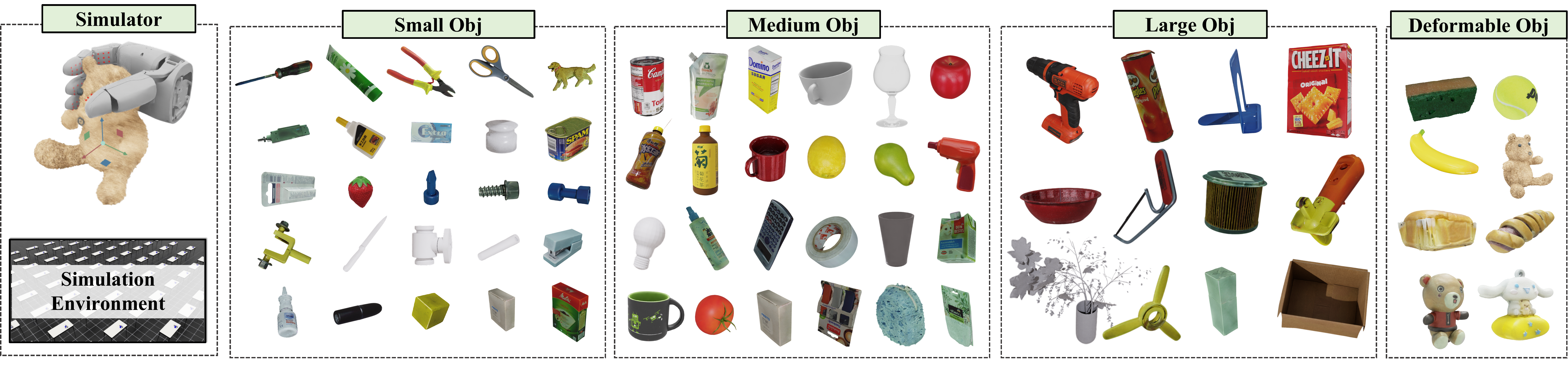}
    \caption{Simulation environment and part of object assets we use in the grasping exeriments.}
    \label{fig:assets}
\end{figure}

\begin{figure}[h]
    \centering
    \begin{subfigure}{\linewidth}
        \centering
        \includegraphics[width=\linewidth]{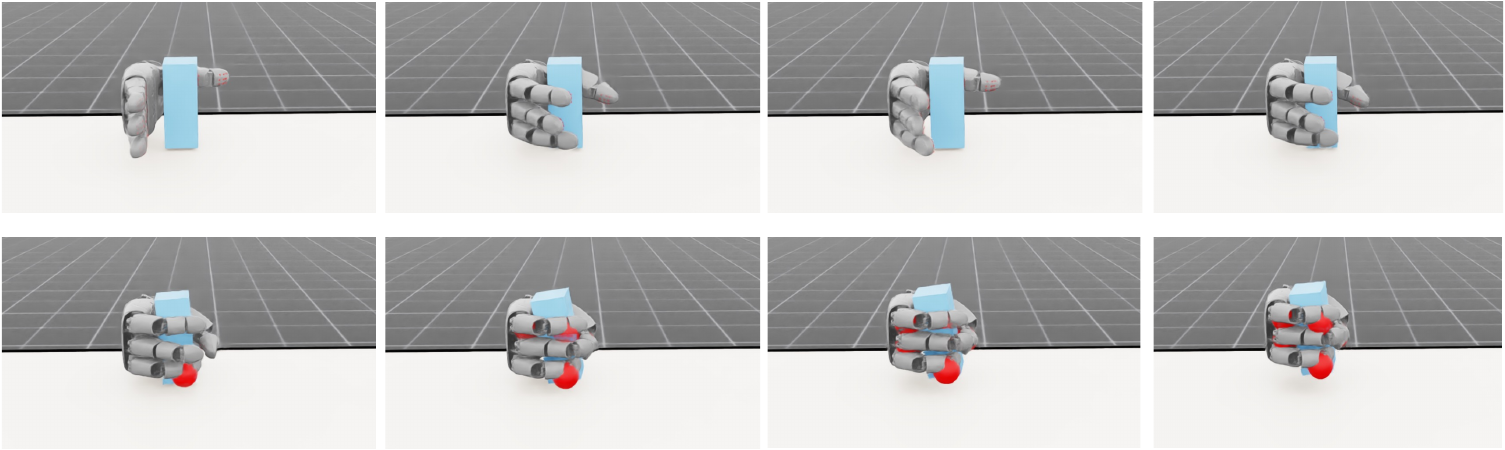}
        \caption{Visualization of cube grasping in the simulation.}
        \label{fig:simulate-cube}
    \end{subfigure}

    \begin{subfigure}{\linewidth}
        \centering
        \includegraphics[width=\linewidth]{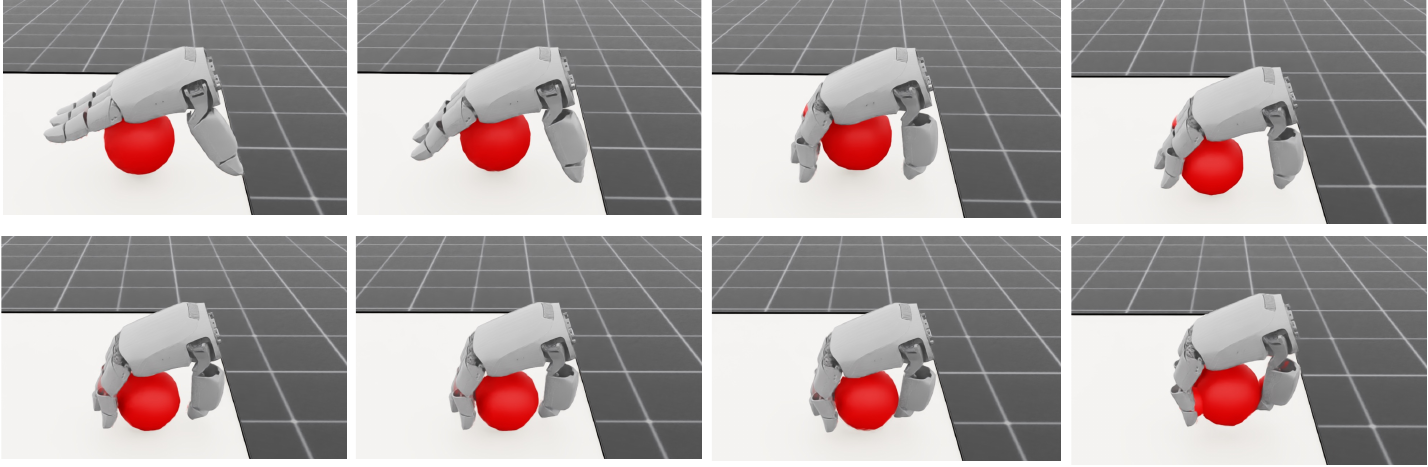}
        \caption{Visualization of ball grasping in the simulation.}
        \label{fig:simulate-ball}
    \end{subfigure}

    \begin{subfigure}{\linewidth}
        \centering
        \includegraphics[width=\linewidth]{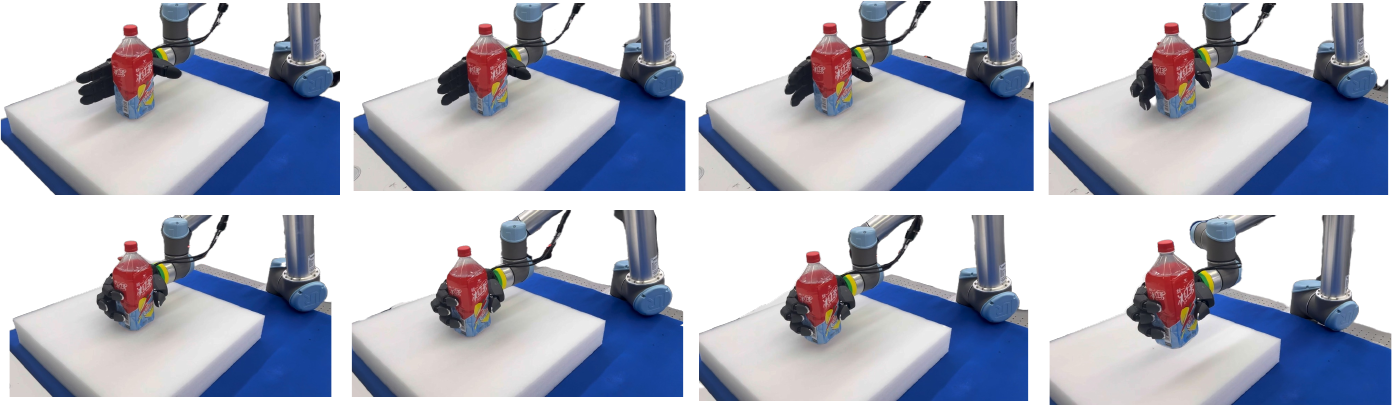}
        \caption{Visualization of ice tea grasping in the real world.}
        \label{fig:real-icetea}
    \end{subfigure}

    \begin{subfigure}{\linewidth}
        \centering
        \includegraphics[width=\linewidth]{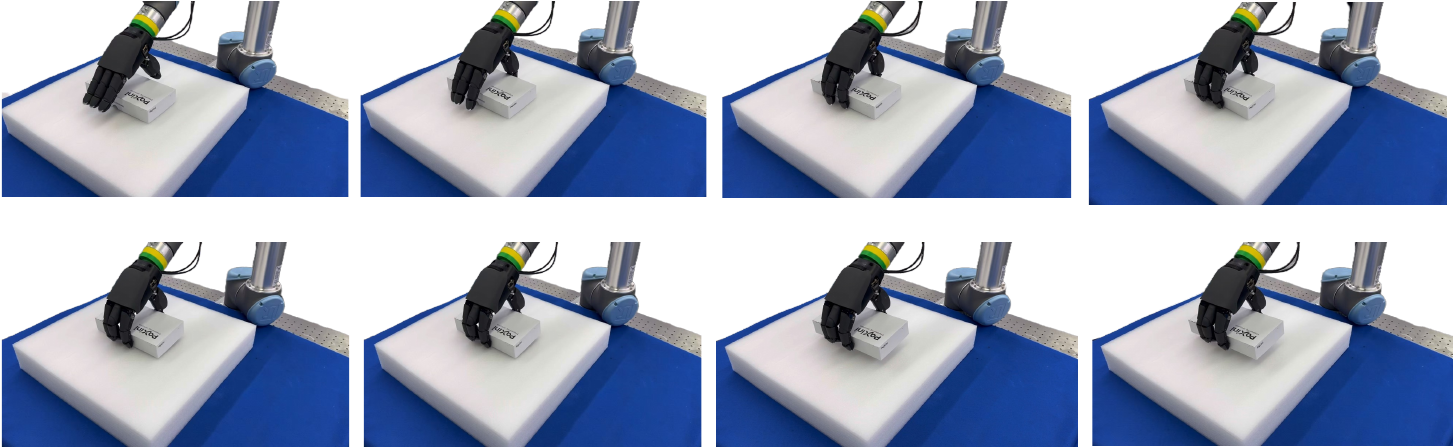}
        \caption{Visualization of box grasping in the real world.}
        \label{fig:real-box}
    \end{subfigure}
    \caption{{Visualization of grasping sequences in the simulation and real-world experiments.}}
    \label{fig:process}
\end{figure}


\subsection{More Lossy Compression Results for Dexterous Grasping}

We further visualize the simulation environment from the IssacSim and part of the assets, as shown in \cref{fig:assets}. The hand model is based on the Paxini DexHand13~\cite{paxinidexhand}, which has four fingers and a total of 16 DoF. Each finger except the thumb is equipped with three tactile sensors and the thumb finger is equipped with two tactile sensors, resulting to a total of 11 tactile sensors. We deploy a simple asymmetric actor-critic (AAC) network with the tactile data as the input, to learn the dexterous grasping for general objects \citep{keyu}. Although the grasping success rate of our baseline model is not very high, we focus on the impact of tactile compression.

We have conducted grasping experiments in real-world settings and employed four mature encoders (JPEG2000, JPEG XL, BPG, VTM) to compress tactile signals with varying quantization parameters (QP). Using four fingertip positions as primary observation metrics, we present the sensory force data along the x, y, and z axes across these four fingertips, with the results illustrated in \cref{fig:process}.

\begin{figure}[h]
    \centering
    \includegraphics[width=\linewidth]{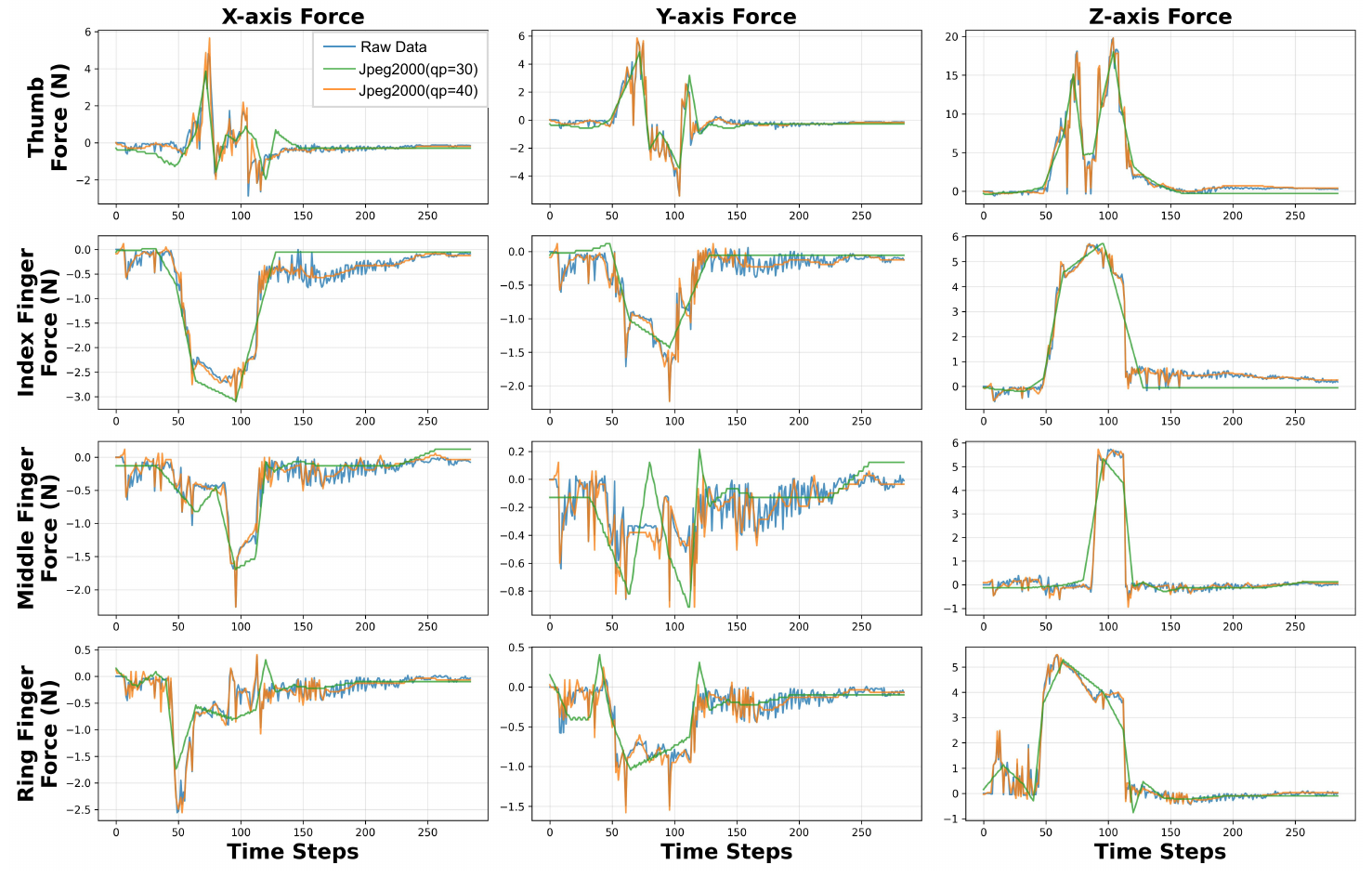}
    \caption{{Visualization of tactile signals in the real-world experiments with JPEG2000 as the codec.}}
    \label{fig:jpeg2000}
\end{figure}

\begin{figure}[h]
    \centering
    \includegraphics[width=\linewidth]{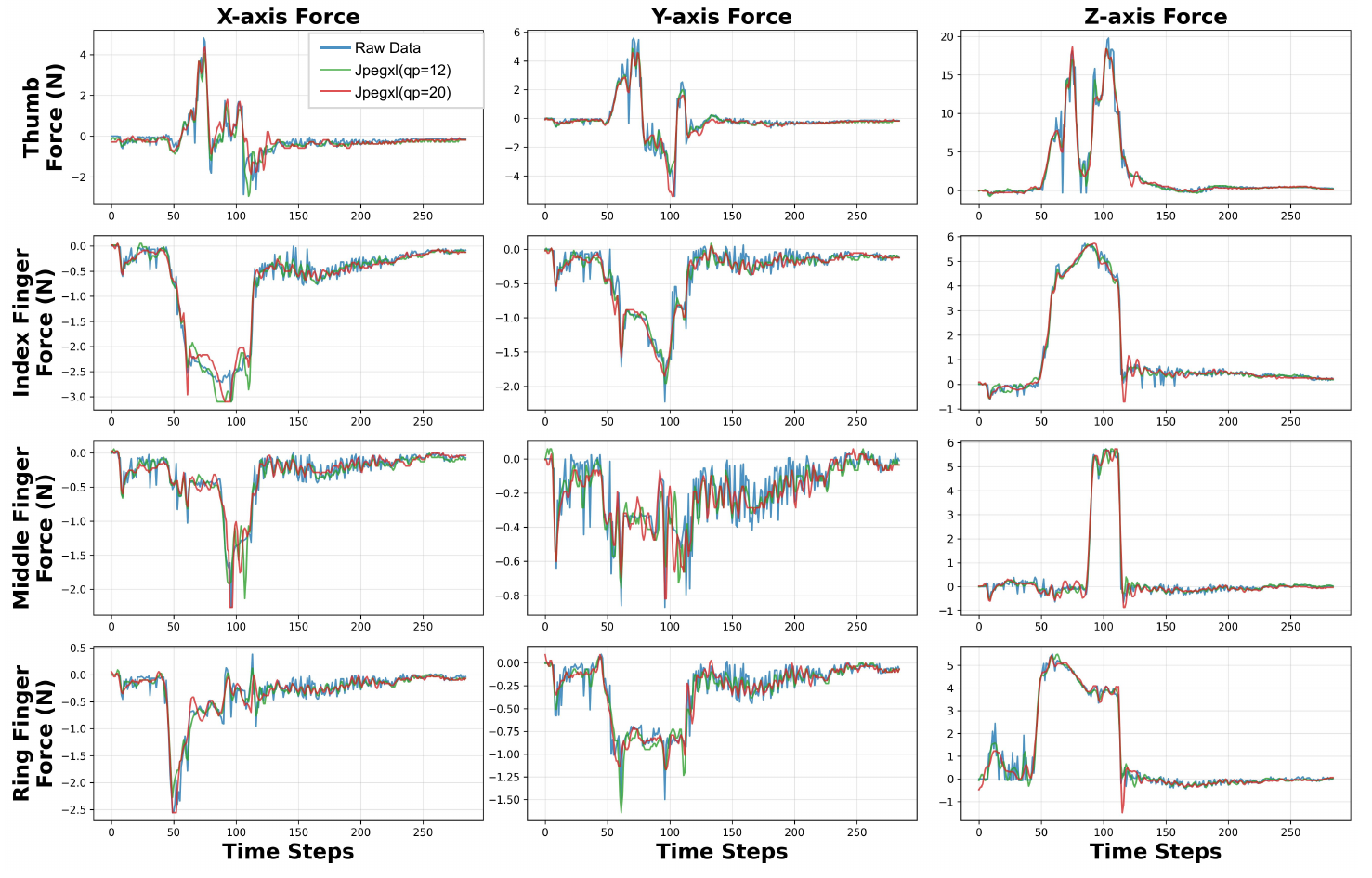  }
    \caption{Visualization of tactile signals in the real-world experiments with JPEG-XL as the codec.}
    \label{fig:jpegxl}
\end{figure}

\begin{figure}[h]
    \centering
    \includegraphics[width=\linewidth]{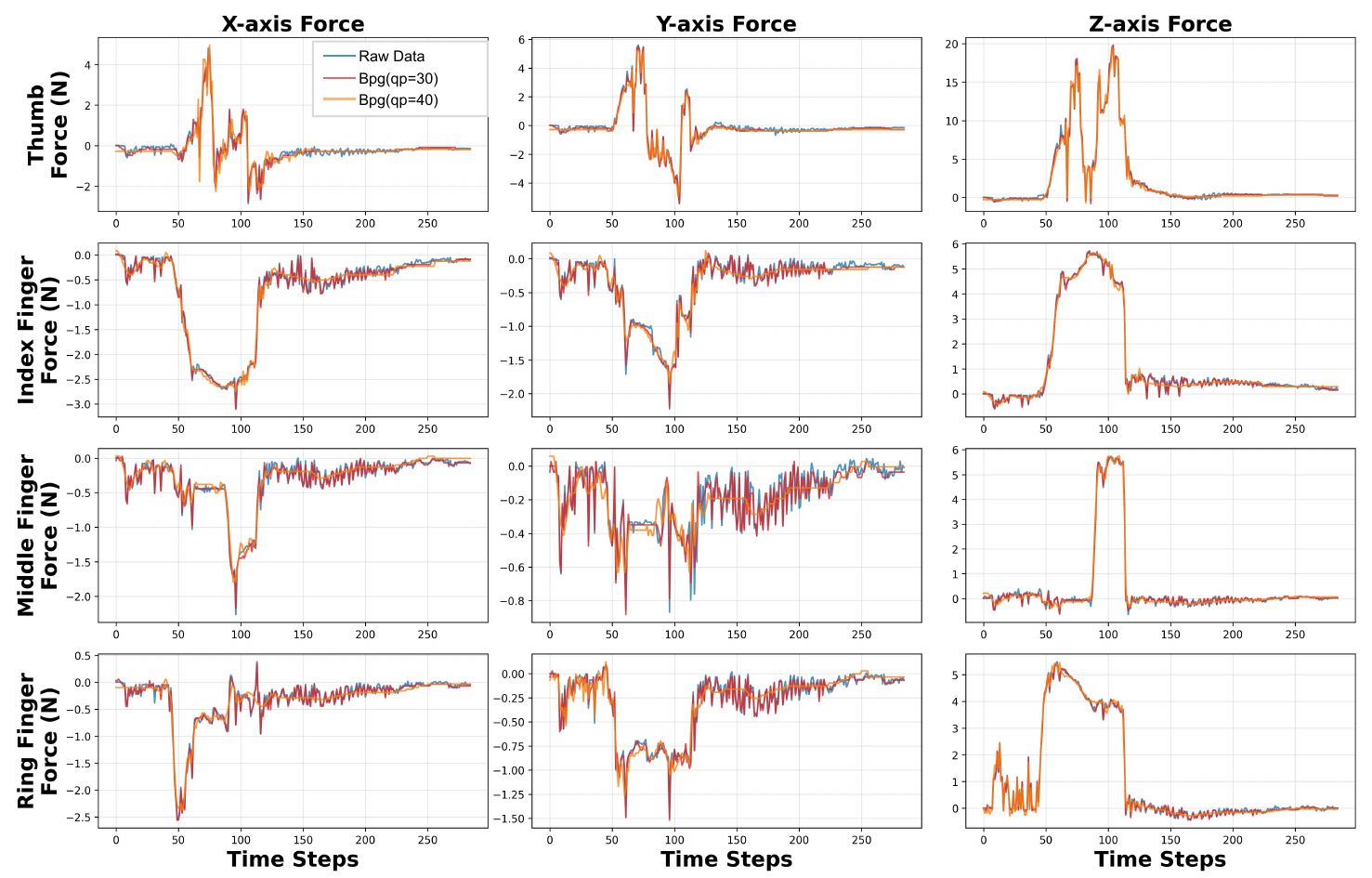  }
    \caption{Visualization of tactile signals in the real-world experiments with BPG as the codec.}
    \label{fig:BPG}
\end{figure}

\begin{figure}[h]
    \centering
    \includegraphics[width=\linewidth]{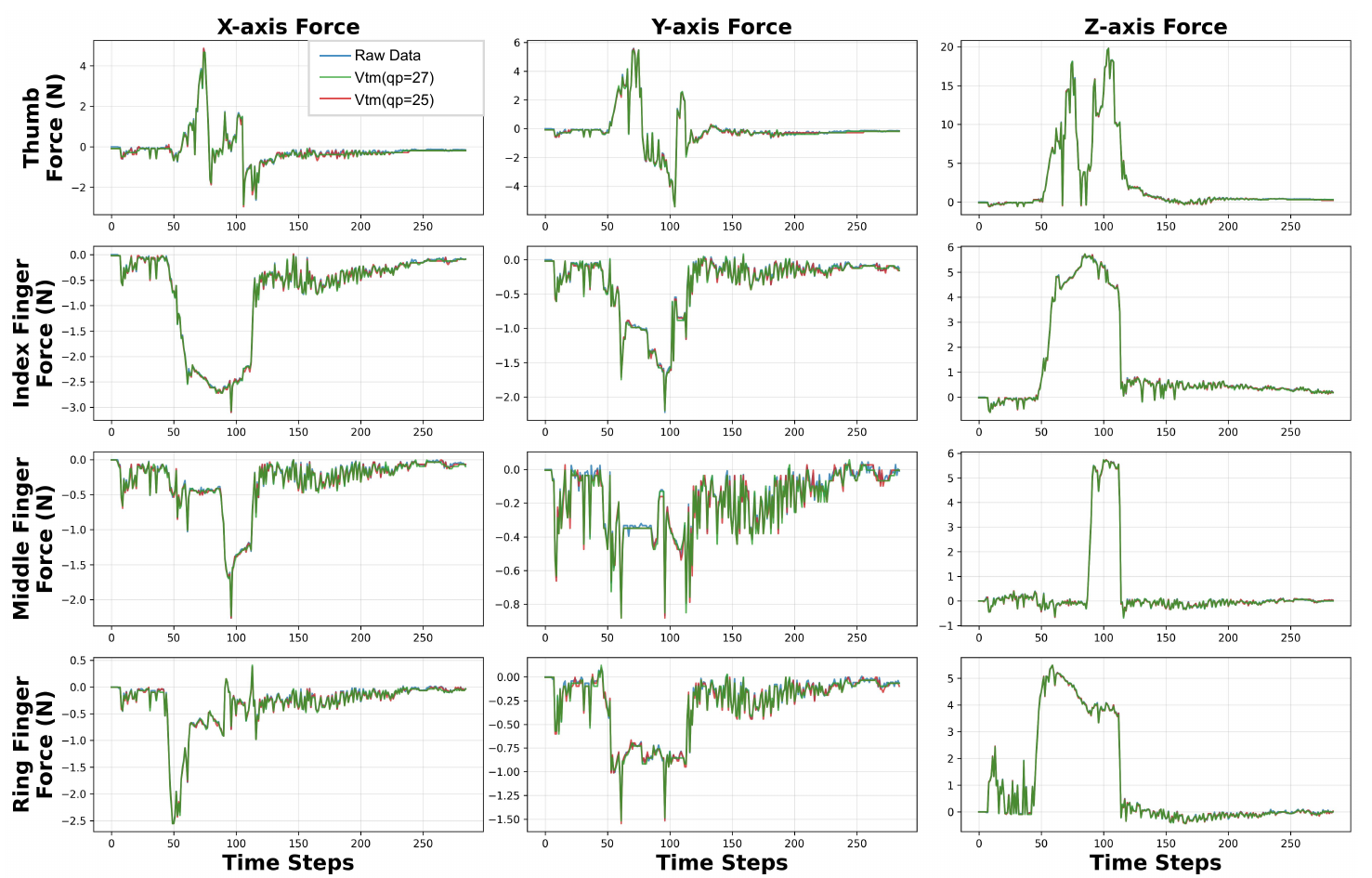 }
    \caption{ Visualization of tactile signals in the  real-world experiments  with VTM as the codec.}
    \label{fig:vtm}
\end{figure}

\begin{figure}[h]
    \centering
    \includegraphics[width=\linewidth]{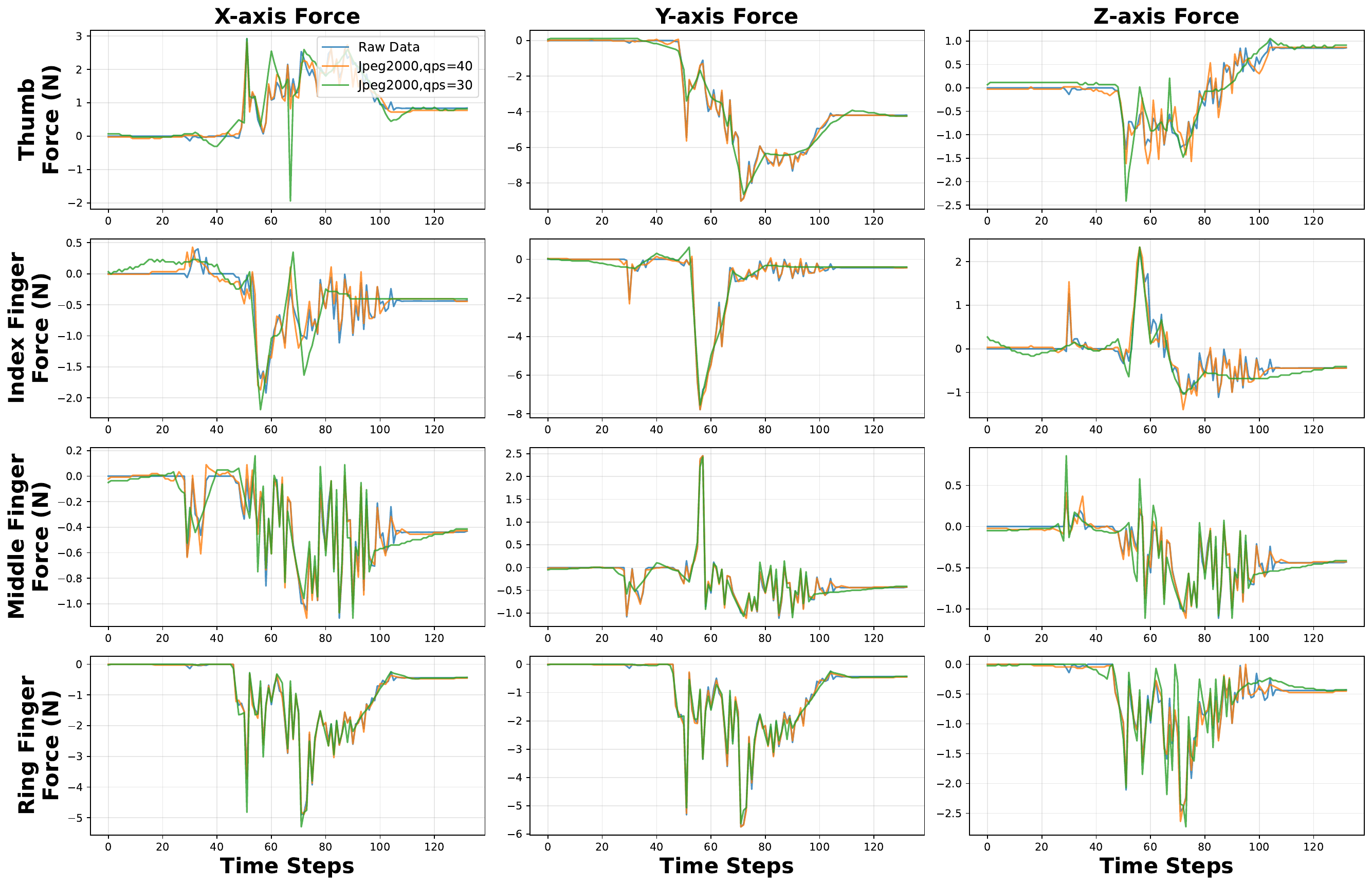}
    \caption{Visualization of tactile signals in the simulation experiments with JPEG2000 as the codec.}
    \label{fig:jpeg2000_ball}
\end{figure}

\begin{figure}[h]
    \centering
    \includegraphics[width=\linewidth]{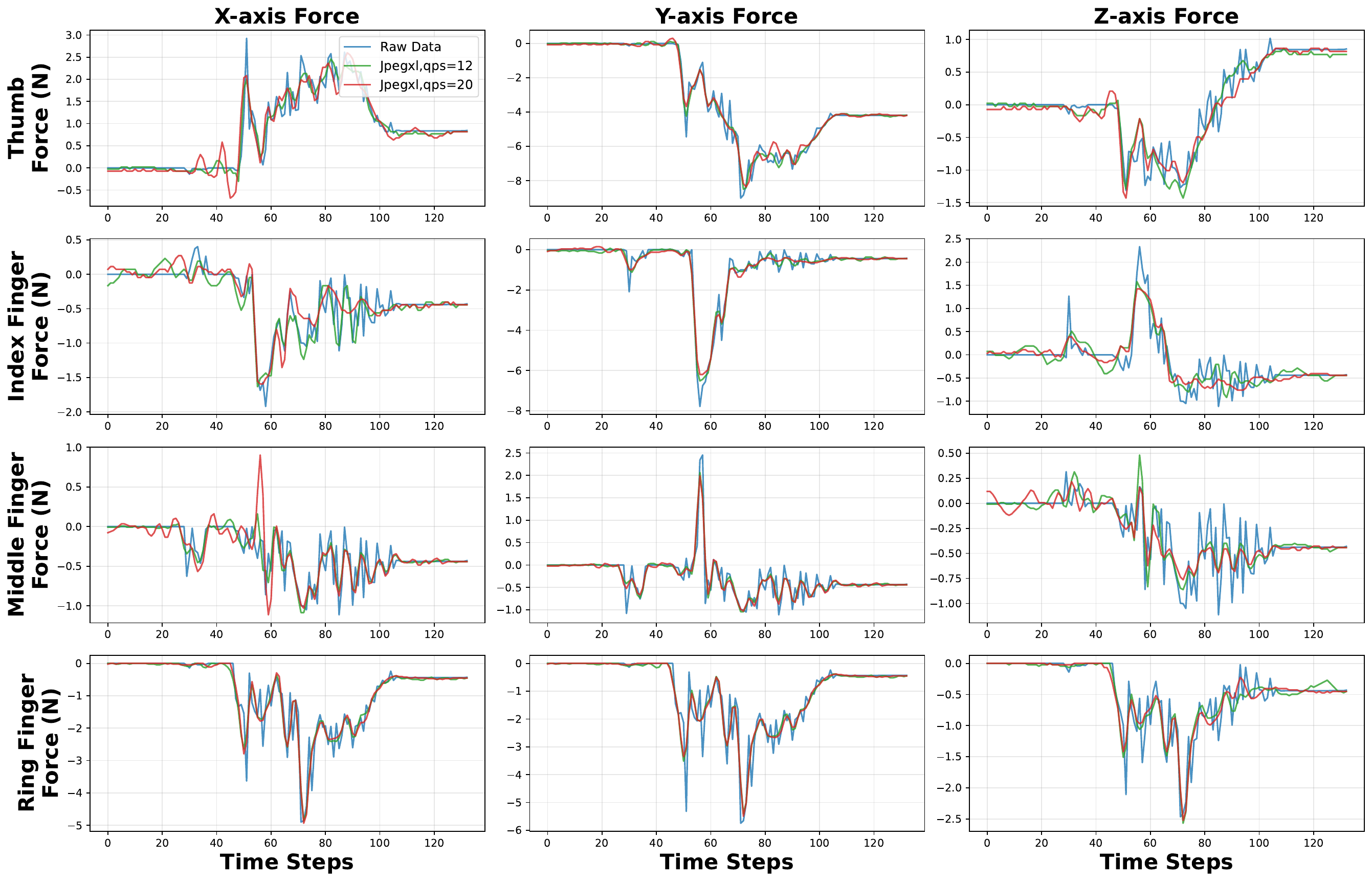  }
    \caption{Visualization of tactile signals in the simulation experiments  with JPEG-XL as the codec.}
    \label{fig:jpegxl_ball}
\end{figure}

\begin{figure}[h]
    \centering
    \includegraphics[width=\linewidth]{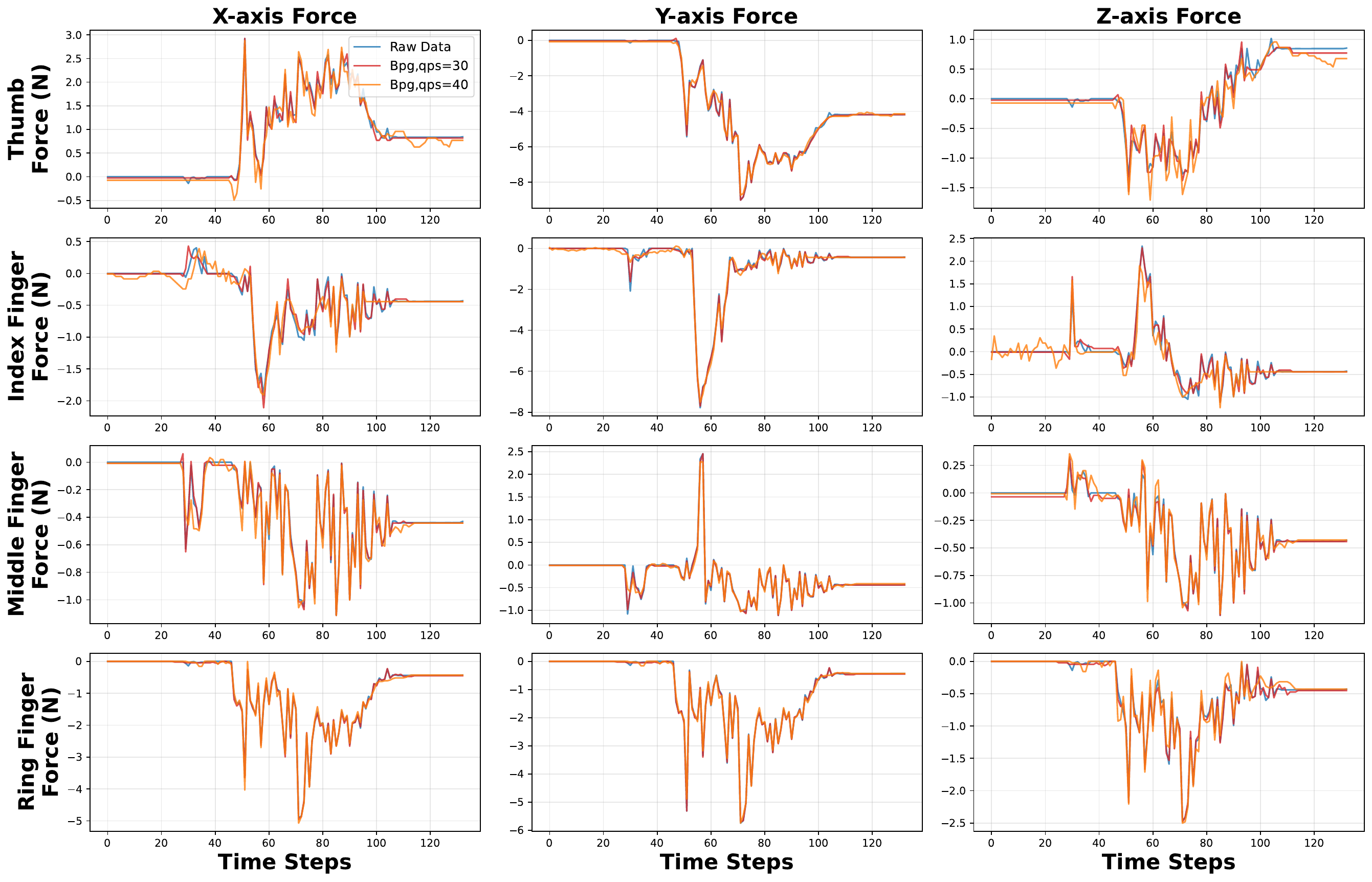  }
    \caption{ Visualization of tactile signals in the simulation experiments  with BPG as the codec.}
    \label{fig:BPG_ball}
\end{figure}

\begin{figure}[h]
    \centering
    \includegraphics[width=\linewidth]{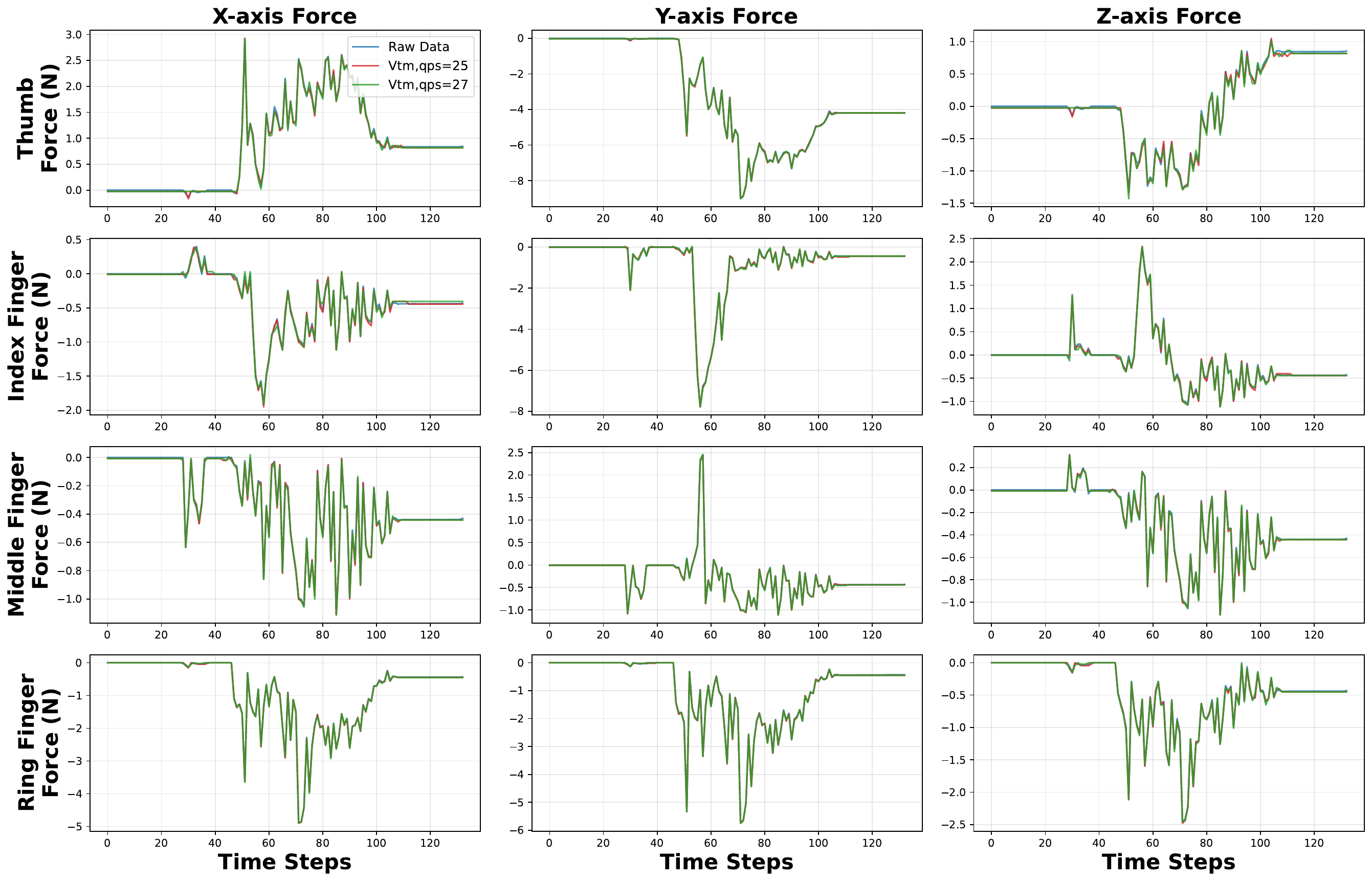 }
    \caption{ Visualization of tactile signals in the simulation experiments  with VTM as the codec.}
    \label{fig:vtm_ball}
\end{figure}

\cref{fig:jpeg2000}, \cref{fig:jpegxl},\cref{fig:BPG} and \cref{fig:vtm} illustrate the visualization results of tactile signals from four fingertips using four different codecs in real-world experiments. Meanwhile, 
\cref{fig:jpeg2000_ball}, \cref{fig:jpegxl_ball},\cref{fig:BPG_ball} and 
\cref{fig:vtm_ball} illustrate the visualization results of tactile signals from four fingertips using four different codecs in the simulations. These figures demonstrate that the compression algorithm itself does not actually affect the main variation distribution of the tactile data, and therefore will not have a catastrophic impact on the accuracy of real-world tasks.

Regarding the implementation details, the simulation environment for the reinforcement learning controller operates at a control frequency of 100 Hz, which is determined by the simulation time step of 0.01 seconds. Specifically, (1) the tactile sensors are updated at every simulation step, resulting in a tactile sampling rate of 100 Hz. (2) The overall latency of the control loop is approximately 0.01 seconds, plus the time required for policy inference. The policy inference is performed using an ONNX model, and the inference time is logged during execution. If the inference time exceeds the simulation time step, the control frequency may decrease, and the latency would increase accordingly. (3) When the combined codec and inference latency approximately equals the simulation update interval, the additional delay introduced to the simulation environment becomes negligible, as it aligns with the natural timing cycle of the control loop. However, in the current implementation, the control command is applied in the same simulation step after inference, so the latency is primarily determined by the simulation step and the inference time.

\subsection{Our Motivation and Future Work}


In this section, we simply present the need for advancing tactile compression. The development of this tactile codec benchmark is motivated by three critical challenges in practical robotics applications. 
First, for dexterous manipulation, tactile data from high-resolution sensor arrays on robot hands can consume a significant portion of the available bandwidth. This is analogous to the Cortical Homunculus, where the hands claim a disproportionately large share of neural resources. The limited bandwidth of low-cost microcontrollers (MCUs) embedded in such hands creates a fundamental bottleneck for real-time sensorimotor control. 

Second, in robotic tele-operation systems, achieving stable and transparent remote control requires low-latency, high-fidelity transmission of tactile signals. Effective compression is paramount to close the feedback loop for delicate tasks, enabling true physical understanding and interaction at a distance. 

Third, to scale up in the field of embodied intelligence, the creation of large-scale training datasets necessitates efficient storage solutions. Specifically, Google introduced Open X-Embodiment Dataset, the largest open-source real robot dataset to date. It contains 1M+ real robot trajectories (download size is \textbf{8965 GB})~\cite{openxdataset}. While video compression is mature, specialized algorithms for tactile data remain underdeveloped, hindering our ability to build and manage the vast datasets required for training generalist robotic models. These pressing needs collectively motivate the establishment of a rigorous benchmark to advance the field of tactile data compression.

For the future work, we will develop a video-like tactile codec by retraining the tactile dataset using the latest neural video compression models, like DCVC-serier models. 


\subsection{LLM Usage Statement}

Large Language Models (LLMs) were not used during the research, experimentation, or analysis phases of this work. During the manuscript preparation, LLMs were used solely for minor grammar and language refinements. No content, ideas, or technical writing was generated by LLMs.

\end{document}